\documentclass[runningheads]{llncs}
\usepackage[T1]{fontenc}
\usepackage{graphicx}
\usepackage{multirow}
\usepackage{caption}
\usepackage{subcaption}  % Include this package for subtables
\usepackage{svg}
\usepackage{hyperref}
\usepackage{todonotes}
\usepackage{float}
 \usepackage{tcolorbox} 
\begin{document}
\title{Are LLMs Truly Multilingual? \\ Exploring Zero-Shot Multilingual Capability of LLMs for Information Retrieval: An Italian Healthcare Use Case}
\titlerunning{Are LLMs Truly Multilingual?}
\author{Vignesh Kumar Kembu \inst{1,2}\orcidID{0009-0002-3782-8111} \and \\
Pierandrea Morandini\inst{2}\orcidID{0000-0002-8615-3766} \and \\
Marta Bianca Maria Ranzini\inst{2}\orcidID{0000-0001-8275-6028} \and \\
Antonino Nocera\inst{1}\orcidID{0000-0003-2120-2341}}
\authorrunning{VK. Kembu et al.}
\institute{Department of Electrical, Computer and Biomedical Engineering,\\ 
University of Pavia, Italy\\
\email{vigneshkumar.kembu01@universitadipavia.it}, \email{antonino.nocera@unipv.it}\\
 \and
IRCCS Humanitas Research Hospital, Milan, Italy\\
\email{\{vignesh.kembu,pierandrea.morandini,marta.ranzini\}@humanitas.it}}
\maketitle              % typeset the header of the contribution
\begin{abstract}
Large Language Models (LLMs) have become a key topic in AI and NLP, transforming sectors like healthcare, finance, education, and marketing by improving customer service, automating tasks, providing insights, improving diagnostics, and personalizing learning experiences. Information extraction from clinical records is a crucial task in digital healthcare. Although traditional NLP techniques have been used for this in the past, they often fall short due to the complexity, variability of clinical language, and high inner semantics in the free clinical text. Recently, Large Language Models (LLMs) have become a powerful tool for better understanding and generating human-like text, making them highly effective in this area. In this paper, we explore the ability of open-source multilingual LLMs to understand EHRs (Electronic Health Records) in Italian and help extract information from them in real-time. Our detailed experimental campaign on comorbidity extraction from EHR reveals that some LLMs struggle in zero-shot, on-premises settings, and others show significant variation in performance, struggling to generalize across various diseases when compared to native pattern matching and manual annotations.

\end{abstract}
\section{Introduction}
Large Language Models (LLMs) have revolutionized the field of natural language processing, showcasing impressive capabilities in text generation, comprehension, and conversational interaction. The models, such as GPT's and Google's Bard etc., are based on advanced neural networks with billions of parameters. They can grasp context, semantics, and intricate language nuances, which allows them to perform exceptionally across diverse applications—from chatbots and virtual assistants to content creation and programming assistance. However, despite their strengths, LLMs encounter several challenges~\cite{radford2019improving,lewis2019bartdenoisingsequencetosequencepretraining}. They can generate incorrect information, may misinterpret subtle input variations and have the potential to produce biased or inappropriate content. Ongoing research seeks to address these issues through improved training techniques, fine-tuning processes and the establishment of ethical guidelines. Today, LLMs play a crucial role in various sectors, including education, healthcare and customer service, offering innovative solutions while also raising important questions about security, privacy and the future of human-computer interactions. As these models continue to evolve, their potential to transform communication and creativity appears huge, presenting both exciting opportunities and complex challenges for society~\cite{10.1145/3641289,zhao2025surveylargelanguagemodels}.

In context of healthcare LLMs are revolutionizing healthcare by enhancing clinical decision-making, automating administrative processes and improving patient engagement. These advanced AI systems, trained on large datasets, can interpret and generate human-like text, making them valuable for various applications in healthcare. One significant area of impact is in clinical decision support. LLMs can analyze medical literature, patient records and research data to provide evidence-based recommendations to healthcare professionals. By synthesizing complex information quickly, they aid clinicians in diagnosing conditions and selecting appropriate treatments, ultimately improving patient outcomes and are streamlining administrative tasks such as scheduling, billing and documentation~\cite{li2024scopingreviewusinglarge,gu2024scalable,thirunavukarasu2023large}. 

\textbf{Scenario.}\label{q1} In a clinical context, we consider a clinician aiming to extract comorbidities from EHRs using LLMs, focusing on a simple and direct approach, which does not need any manipulation of the LLM prompt(Zero-shot). Considering this, we formulate three key questions to be addressed in the evaluation of the scenario.
\begin{itemize}
    \item \textbf{Q1.}\label{q2} Can we use LLMs in Zero-shot to extract comorbidities from EHRs?
    \item \textbf{Q2.}\label{q3} Can LLMs substitute a regular expression-based approach? 
    \item \textbf{Q3.}\label{q4} Can we find a best LLM among the chosen ones?
\end{itemize}

% Patient engagement is another critical application. LLMs can power chatbots that provide immediate responses to patient queries, offer medication reminders and assist in symptom evaluation. These tools enhance patient experience and adherence to treatment protocols, fostering a more interactive healthcare environment. The integration of LLMs in healthcare is not without challenges. Ensuring data privacy, managing biases and validating the accuracy of AI-generated recommendations are crucial issues that must be addressed. Additionally, the ethical implications of relying on AI for clinical decision-making need careful consideration. LLMs represent a transformative force in healthcare, offering promising advancements in clinical support, administration and patient interaction. As the technology continues to evolve, its responsible implementation will be essential for maximizing benefits while minimizing risks. 

Extending from the scenario, we evaluate LLMs (Large Language Models) in multilingual settings within digital health and clinical decision support.  This research explores the use of different LLMs in the healthcare domain, specifically focusing on the classification tasks involving Italian patient EHRs. The objective was to design, develop a generalized data gathering ETL of patient electronic health records, LLM pipeline and framework which can handle different data formats and ranges of models with different parameters, capable of extracting diverse types of information from clinical records. In context with this, our current study leverages usage of 6 open-source models from three model families. The inference was carried out in Zero-shot setting on free EHR text of patients, which are discussed below in detail.

\section{Preliminaries}
% In this section, we discuss large language models and examine their multilingual abilities for information retrieval(IR).

\textbf{Large Language Models (LLMs)} are advanced models designed to process and generate human language, trained on vast amount of text. These models use network architecture called transformers~\cite{vaswani2023attentionneed}, which help them manage and produce text as in human communication. These models are typically built with billions of parameters and more, allowing them to capture internal patterns in language, context and reasoning~\cite{kaplan2020scalinglawsneurallanguage}. 
% The ability of these models to transfer knowledge across different tasks has made them valuable tools in industries such as healthcare, finance and entertainment, where they can assist in decision-making, drafting reports and  coding~\cite{radford2019improving,lewis2019bartdenoisingsequencetosequencepretraining}.
% LLMs support by summarizing records, analyzing medical literature, and giving evidence-based recommendations. They also improve patient care through virtual assistants for queries, mental health support, and follow-ups~\cite{chowdhery2022palmscalinglanguagemodeling}.

$Closed$ source models, such as GPT~\cite{brown2020languagemodelsfewshotlearners,openai2024gpt4technicalreport}, Gemini~\cite{team2023gemini,comanici2025gemini}, and Claude~\cite{claude3,claude4}, are proprietary and accessible via APIs, enabling businesses to integrate advanced AI without building models from scratch. But  their closed nature raises security concerns, particularly in sensitive areas like healthcare. On the other hand, $Open$ source models like Qwen's~\cite{qwen2.5,yang2025qwen3technicalreport}, Bloom's~\cite{workshop2022bloom}, Llama's~\cite{llama1,llama2}, and Mistral's~\cite{jiang2023mistral7b,jiang2024mixtralexperts}, offer transparency and customization, allowing fine-tuning for specific tasks, such as healthcare applications, while providing better control over data privacy and regulatory compliance.

% $Closed$ source refer to models that are proprietary and not open to the public for inspection or modification, which are accessible via Application Programming Interface (API). Closed source LLMs such as GPT's \cite{brown2020languagemodelsfewshotlearners,openai2024gpt4technicalreport}, Gemini's~\cite{team2023gemini,comanici2025gemini}, Claude's~\cite{claude3,claude4} and more enable businesses to add advanced AI features without building models from scratch. But the proprietary nature of these models raises security concerns, notably regarding data privacy in sensitive areas like healthcare. $Open$ source refer to models whose architecture and weights are freely available. These models provide transparency, flexibility and control over how they can be deployed (on-premises), allowing researchers and organizations to customize them for specific use cases, such     as domain based finetuning ~\cite{wolf2020huggingfacestransformersstateoftheartnatural}. Few notable models such as Qwen's~\cite{qwen2.5,yang2025qwen3technicalreport}, Bloom~\cite{workshop2022bloom}, Llama's~\cite{llama1,llama2},  mistral's~\cite{jiang2023mistral7b,jiang2024mixtralexperts} and more provide flexibility to fine-tune the models for specific tasks or domains. In healthcare, open-source models enable special applications like medical chatbots and clinical decision support, while offering greater control over data security and privacy, essential for regulatory compliance.

\textbf{Information Extraction (IE)} is a crucial task in natural language processing (NLP) that involves automatically extracting structured information from unstructured text. This process is key for converting large volumes of textual data, such as news articles, medical text (EHRs) and social media posts, into usable information for further analysis or decision-making\cite{Nadeau2007ASO}. Large Language Models (LLMs), such as GPT and BERT, have significantly advanced the field of Information Extraction (IE). These models, trained on massive corpora, excel at identifying entities, relationships and other structured information in diverse and unstructured text without requiring task-specific training data \cite{devlin-etal-2019-bert}. By leveraging their deep contextual understanding, LLMs have shown superior performance in extracting nuanced and complex information, enabling more accurate and adaptable IE systems across various domains \cite{brown2020languagemodelsfewshotlearners}.

\textbf{Multilingual} understanding and generation have made notable progress through models trained on large and diverse multilingual data combined with advanced training techniques. Large language models demonstrate impressive robustness in English, leveraging abundant data and resources. Evidence on their performance and reliability in other languages remains limited and underexplored~\cite{QIN2025101118}. LLMs show potential in healthcare, helping in medical Q\&A, diagnosis, counseling, and EHR data extraction, even in multilingual settings~\cite {zhu2024multilinguallargelanguagemodels}.

\textbf{Electronic Health Records (EHRs)} are digital information that contains patients medical history, such as diagnoses, treatments, medications, laboratory results and documentation notes from clinicians. EHRs improve continuity of care and efficiency by allowing the update of patient information in real time~\cite{hayrinen2008definition}. However, most EHR data is unstructured, making it difficult to extract meaningful insights. Information extraction on EHR data was traditionally addressed by regular expressions, however building the pattern is complicated, not generalizable and requires deep domain knowledge with all linguistics nuances of the domain. Currently, Large Language Models (LLMs) have great potential to process both structured and unstructured data, enabling them to extract valuable insights for clinical decision-making and research~\cite{li2024scopingreviewusinglarge}. LLMs can extract data from clinical notes with high accuracy, outperforming traditional pattern-matching methods. They are particularly effective in retrieving data, such as lab results and vital signs, which are crucial for clinical analysis.
~\cite{gu2024scalable}. 

\section{Methodology}\label{methodology}
\begin{figure}[t]
    \centering
    \includegraphics[width=0.8\linewidth]{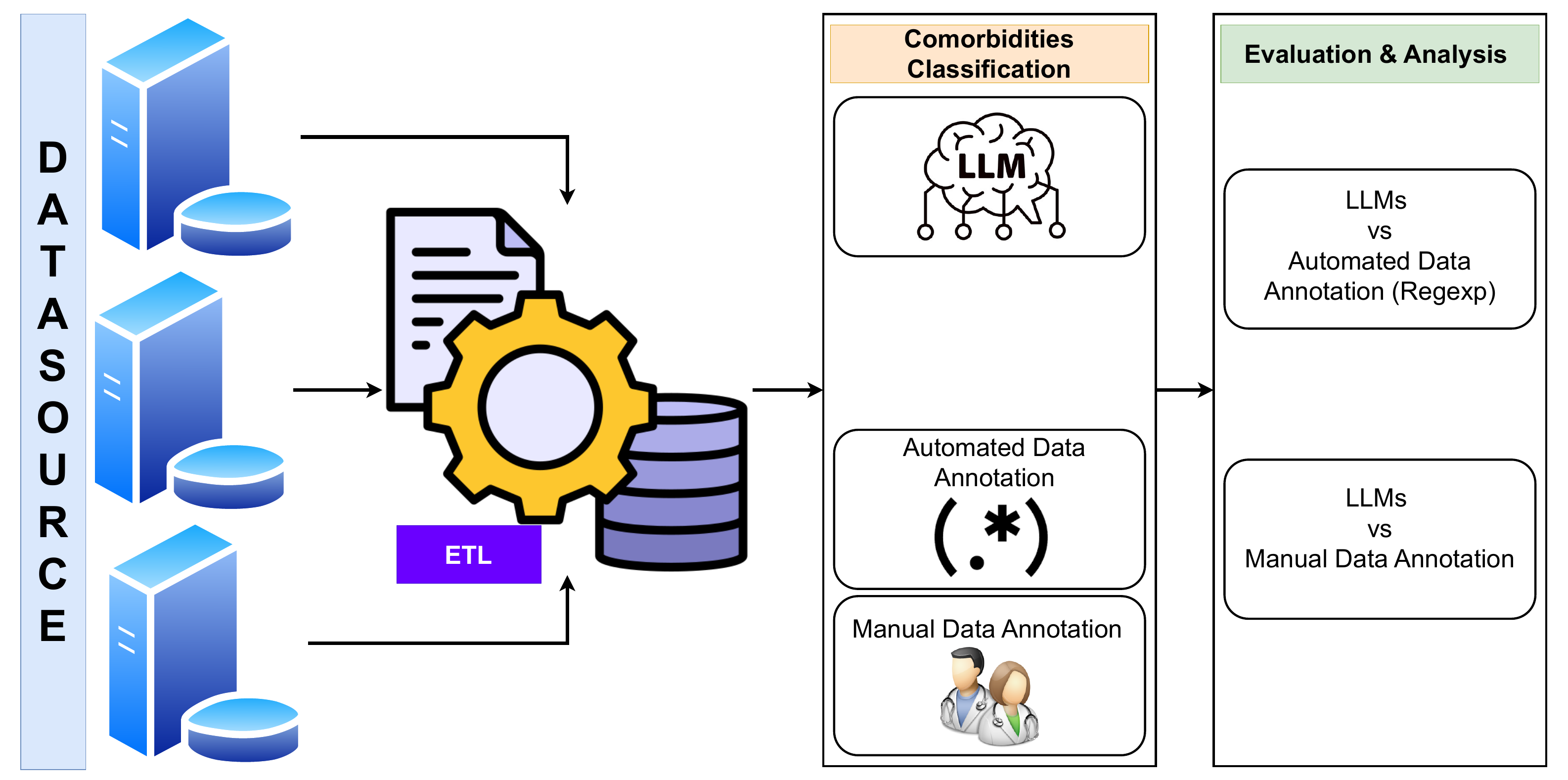}
    \caption{Methodology - Data gathering pipeline, data annotation and comparison: from regexp based automatic classification and clinicians validated ground truth to LLM extraction.}
    \label{fig:exp_setup}
\end{figure}

The methodology comprises of EHR data gathering pipeline, first we applied regular expressions to automatically annotate the texts, establishing a measure for comparison. Then we leveraged a LLMs to extract comorbidities from the same set of EHRs, comparing the LLMs outputs to the Regexp annotations. To manage inaccuracies in the regexp annotations, we chose 100 regexp-false classified EHRs and proceeded with manual annotation by clinicians, creating a ground truth. Finally, further comparison of the LLMs performance against these manually annotated labels was carried out. Figure~\ref{fig:exp_setup} presents the layout of the methodological approach.
\subsection{Data Pipeline \& description}
The protocol required collection of different data from the patient EHR from different hospital areas.
For this, a large ETL using Oracle and Python has been implemented for the collection of the cardiological risk factors, presence of previous cardiac interventions, treatment received during the hospitalization, labs, discharge diagnosis, at-home treatment, and different dataframes have been created: Anamnesis, Interventions, Labs, Procedures, Procedures ICD9, Therapies. We will focus solely on \emph{Anamnesis} data which has been obtained by performing a large ETL process because it represents unstructured text with the comorbidity information of interest. The dataframe consists of three columns, which are: Nosologico (hospital admission ID), Ente (hospital identifier), Anamnesis (EHR free text). With a total of 8223 patient records.

To emphasize the IR part, we focused on the following comorbidities which are clinically relevant for patient evaluation in the cardiac domain: Fibrillazione atriale (Atrial fibrillation), Insufficienza Renale (Kidney failure), BPCO - Broncopneumopatia cronica ostruttiva (COPD-Chronic obstructive pulmonary disease), Diabete mellito (Diabetes mellitus) and Ipertensione arteriosa (Hypertension) will be the key focus of our research. Since the EHR is in the Italian language, all the keys shall be used as per it.

\subsection{Automated Data Annotation using Regular expression}
Building the regexp for the key comorbidities was done in collaboration with the help of clinicians, since the electronic health records are created and updated by them for each patient. So, to create the best regexp classifier, domain experts have been involved in developing the patterns. An example of why we need domain experts for this regexp creation is as follows, Different clinical mentions for the same term: Term: ``Diabetes” - Diabetes mellitus - refers to the general disease entity, DM - Abbreviation for Diabetes Mellitus, often used in medical notes and prescriptions, Type 1 diabetes mellitus, Type 2 diabetes mellitus, Insulin-dependent diabetes mellitus (IDDM), Non-insulin-dependent diabetes mellitus (NIDDM), Diabetes mellitus with nephropathy, Diabetes mellitus with retinopathy, Diabetes mellitus with neuropathy,  Gestational diabetes mellitus (GDM). A domain expert is essential for identifying key comorbidities and creating accurate regex patterns. These patterns are then applied to EHRs to classify comorbidities, producing baseline data for comparison with LLM results.

\subsection{Manual Data Annotation}
To be more precise on the created pattern for data classification using Automated Data Annotation using Regular expression with the help of clinicians, 100 ``false" classified records (i.e., regexp classified as negative(0) for the presence of the comorbidities) of the patient on all five key comorbidities have been manually annotated with help of clinicians (Doctors). Although EHRs aim to improve patient care, they also pose several challenges. One major concern is data quality, as EHRs are only as good as the data entered by clinicians, which can be prone to errors and inaccuracies. EHRs can lack standardization, resulting in varying formats and terminology across different systems and organizations, making it challenging for healthcare providers to share information and coordinate care. Considering this, the regexp created with the help of clinicians information might not be able to capture all the patterns of the comorbidities. So, manual annotation was carried out to double-check the ``false" classified records of the patient.
In the process of manual annotations, two clinicians annotated all five key comorbidities for each selected EHR. They annotated all five key comorbidities discussing case by case to reach a agreement of the class.

\section{Experimental Setup}
There are various factors to consider when designing the research setup. For our experiment, the primary criteria we have considered are discussed below.

\textbf{Privacy concerns \& Licensing.}
Data privacy is a concern with Large Language Models (LLMs), and organizations must address these issues when deploying them.
Humanitas AI Center and Humanitas Research Hospital prioritize data privacy, healthcare patient data should be securely managed and used for research only on-premises. Recalling the difference between ``Closed-source'' and  ``Open-Source'' models, only ``Open-Source'' ones have been selected as they allow for deployment in the ``on-premises'' environment.

\textbf{Language Support.}
Language support in large language models (LLMs) refers to the number of languages a model can understand and generate text, as well as its proficiency in those languages. The models capabilities can vary significantly depending on their training data, architecture and intended use cases. They can be broadly classified as ``Multilingual Models'' and ``Language-Specific Models''. Multilingual models are explicitly designed to handle multiple languages and often support dozens or even more languages. In our case, patients EHR is in the Italian language, thus only "Multilingual Models" have been chosen and the experiments have been carried out.

\textbf{Size and Resource Requirements.}
The size and resource requirements of LLMs significantly influence their accessibility and usability. Large models typically yield impressive capabilities but come with high demands for computational power, memory and storage, making them more costly and complex to operate. Prior understanding of these requirements was crucial for us to adopt LLMs for the proposed information classification task. High-Performance Computing (HPC) with Multiple GPUs was used to carry out the proposed research.

\textbf{Selected LLMs.}
Considering the points stated before, chosen models show great level of accuracy in the leaderboard~\cite{EuroLingua}, especially in consideration with the Italian language score. From OpenLLaMA family 3B \& 7B models, from Mistral family 7B \& 8x7B models and from Qwen2.5 family 3B \& 7B models have been selected for this study.

\section{Result Analysis}
This section examines and compares the performance of cutting-edge large language models (LLMs) towards regular expression and human annotation in the IR task discussed in section~\ref{methodology}.
\subsection{Performance Comparison: LLMs vs Regexp}
To evaluate the performance of various large language models (LLMs), it is essential to establish a reference dataset for comparison. With the availability of annotated data through automated methods, this dataset can serve as the reference. Consequently, the results generated by different LLMs can be compared against these actual values. 
\begin{figure}[h]
    \centering
    \begin{subfigure}[b]{0.19\textwidth}
        \includegraphics[width=\linewidth]{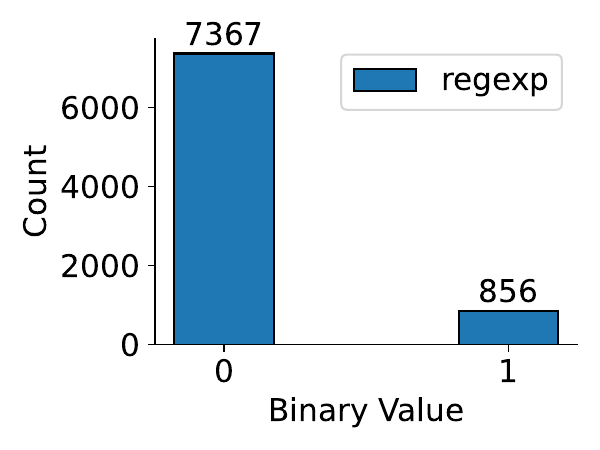}  
        \caption{}
    \end{subfigure}
    \begin{subfigure}[b]{0.19\textwidth}
        \includegraphics[width=\linewidth]{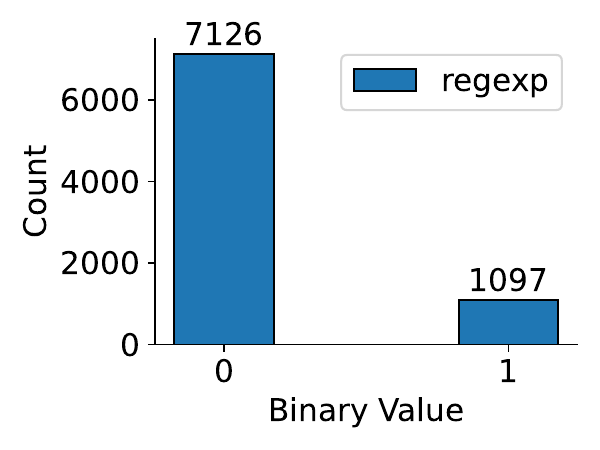}  
        \caption{}
    \end{subfigure}
    \begin{subfigure}[b]{0.19\textwidth}
        \includegraphics[width=\linewidth]{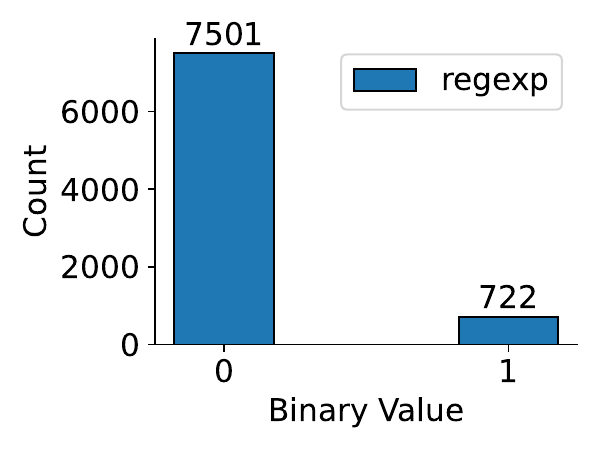}  
        \caption{}
    \end{subfigure}
    \begin{subfigure}[b]{0.19\textwidth}
        \includegraphics[width=\linewidth]{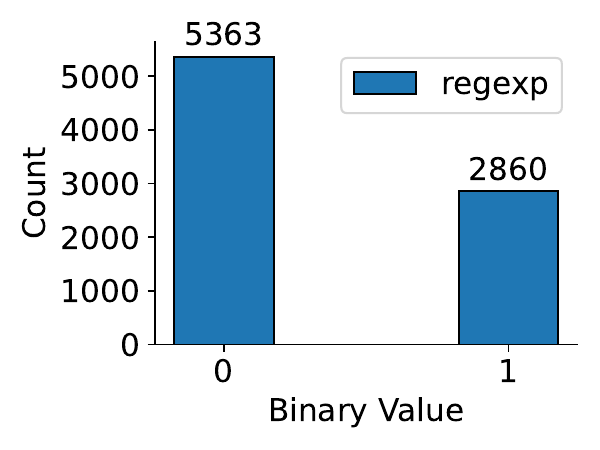}  
        \caption{}
    \end{subfigure}
    \begin{subfigure}[b]{0.19\textwidth}
        \includegraphics[width=\linewidth]{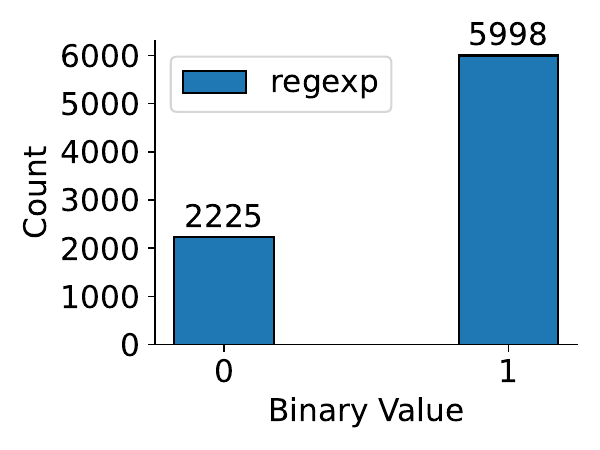}  
        \caption{}
    \end{subfigure}
    \caption{Classification using regular expressions for the chosen comorbidities - a)Fibrillazione atriale, b)Insufficienza Renale, c)BPCO-Broncopneumopatia cronica ostruttiva, d)Diabete mellito and e)Ipertensione arteriosa.}\label{fig:regexp_anno}
\end{figure}
The automated annotation is carried out by applying a series of different combinations of match patterns for the discussed comorbidities, utilizing regular expressions. From the Figure~\ref{fig:regexp_anno} we could see the classification, in which $0$ class represents that the comorbidities is not found in the EHRs and $1$ class represents the availability of the comorbidities in the EHRs. Diabete mellito and ipertensione arteriosa are most positive classified field by the regexp, comorbidities like Fibrillazione atriale, Insufficienza Renale and BPCO were the most negative classified.

\textbf{LLMs.} A standard prompt in a zero-shot setting has been used across all the LLMs in context with the scenario~\ref{q1}. To avoid multiple classifications in a single inference, each comorbidity has been classified in an individual inference for each EHR. Assigning one comorbidity per inference per task maximizes the accuracy of large language models (LLMs) and leads to more precise predictions. This will avoid the misinterpretation of data, especially medical information, which is prone to bias when several tasks are handled by one inference. In executing one task at a time, the model is able to give complete attention to that one task, thereby optimizing the performance. In the following, the classification results of different LLM family models against the regular expressions have been discussed.

\begin{figure}[h]
    \centering
    \begin{subfigure}[b]{0.3\textwidth}
        \includegraphics[width=\linewidth]{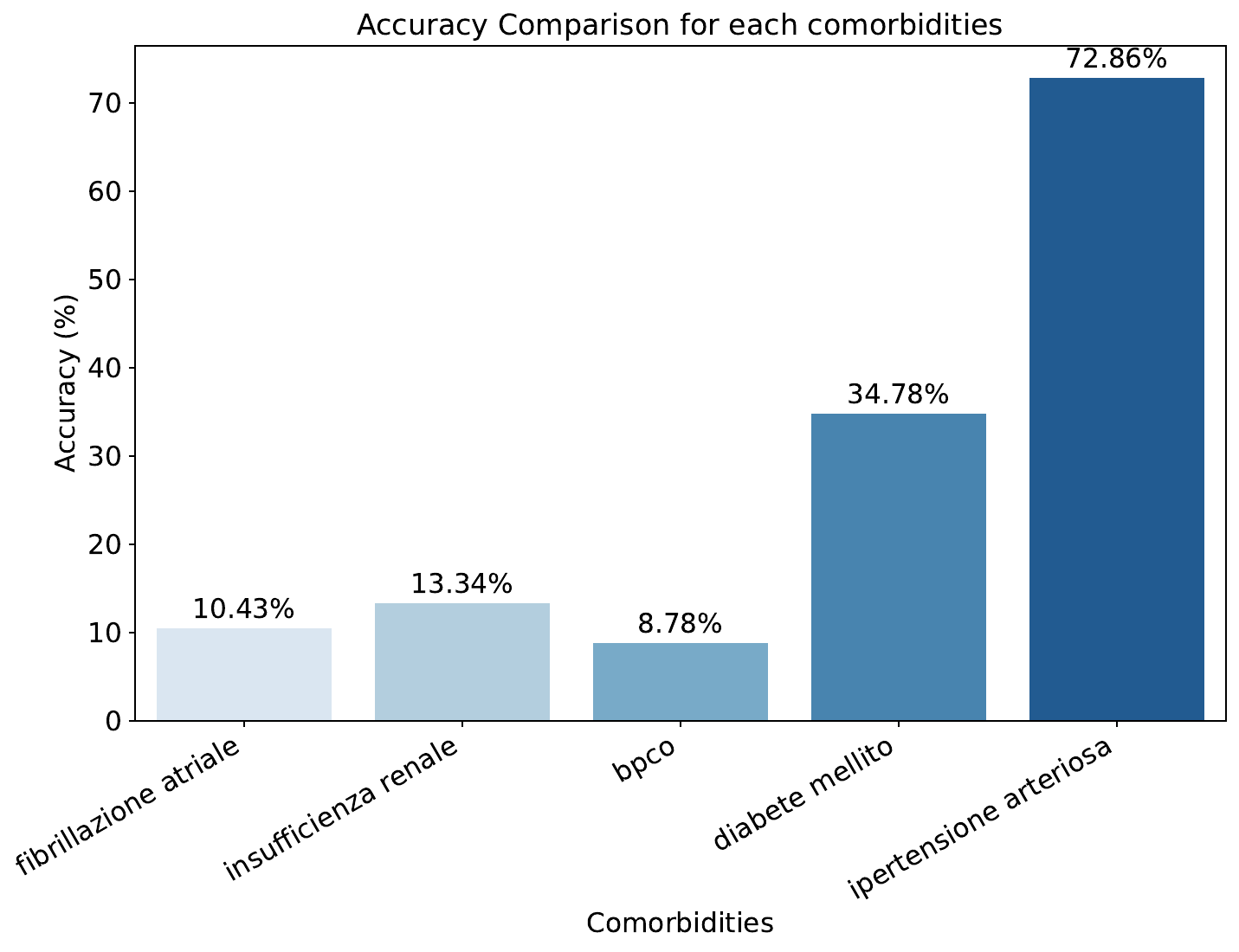}  
        \caption{}
    \end{subfigure}
    \begin{subfigure}[b]{0.3\textwidth}
        \includegraphics[width=\linewidth]{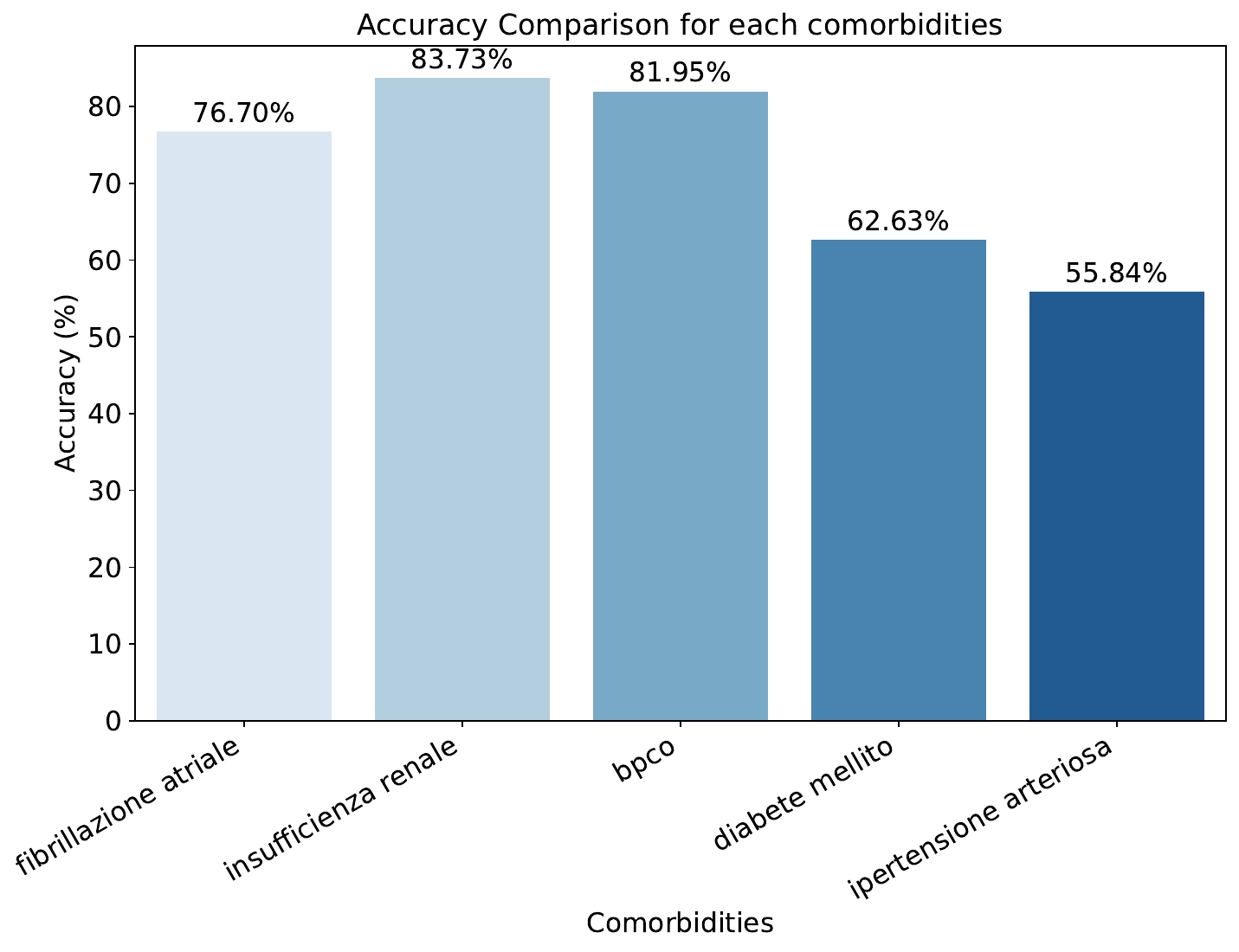}  
        \caption{}
    \end{subfigure}
    \begin{subfigure}[b]{0.3\textwidth}
        \includegraphics[width=\linewidth]{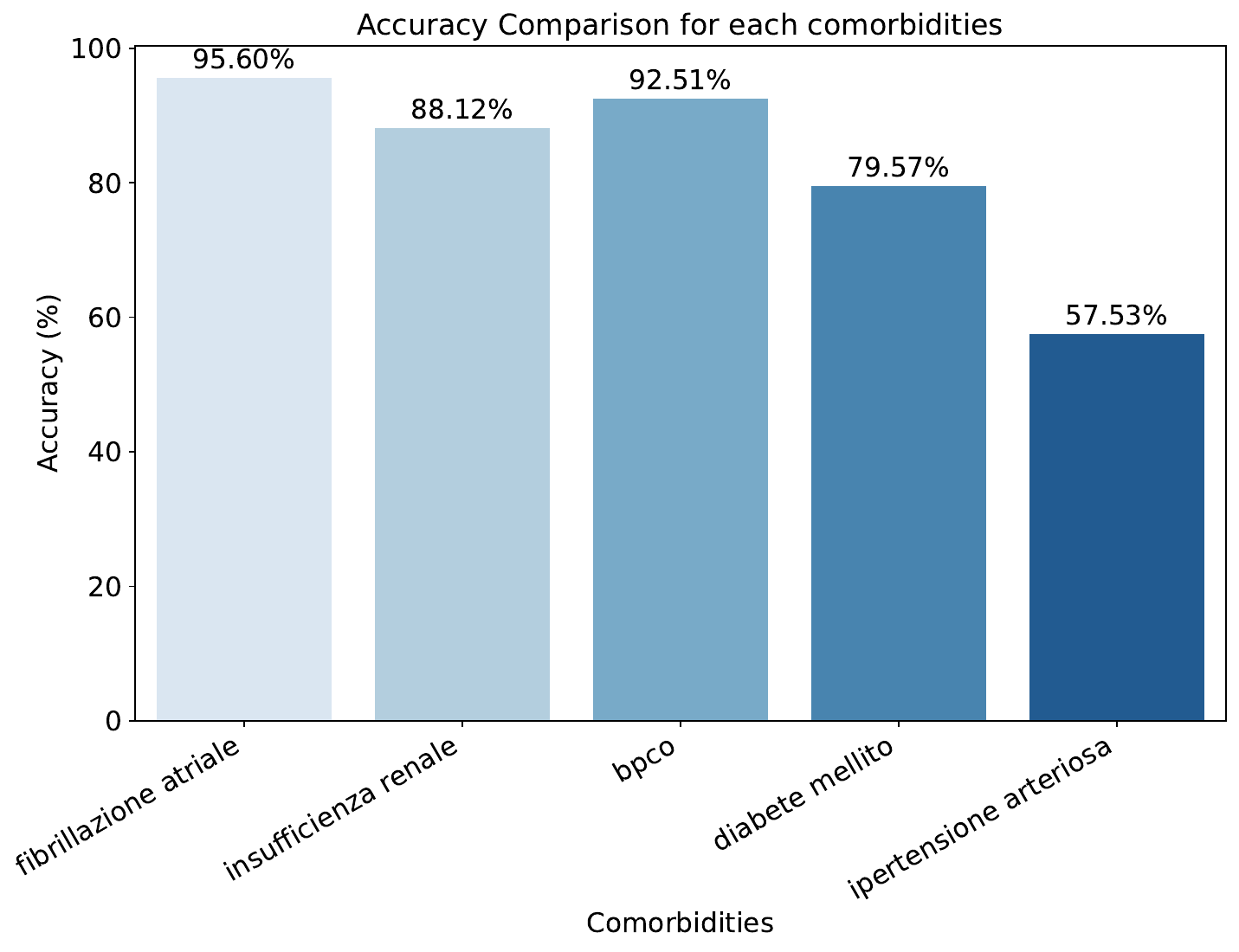}  
        \caption{}
    \end{subfigure}
    \begin{subfigure}[b]{0.3\textwidth}
        \includegraphics[width=\linewidth]{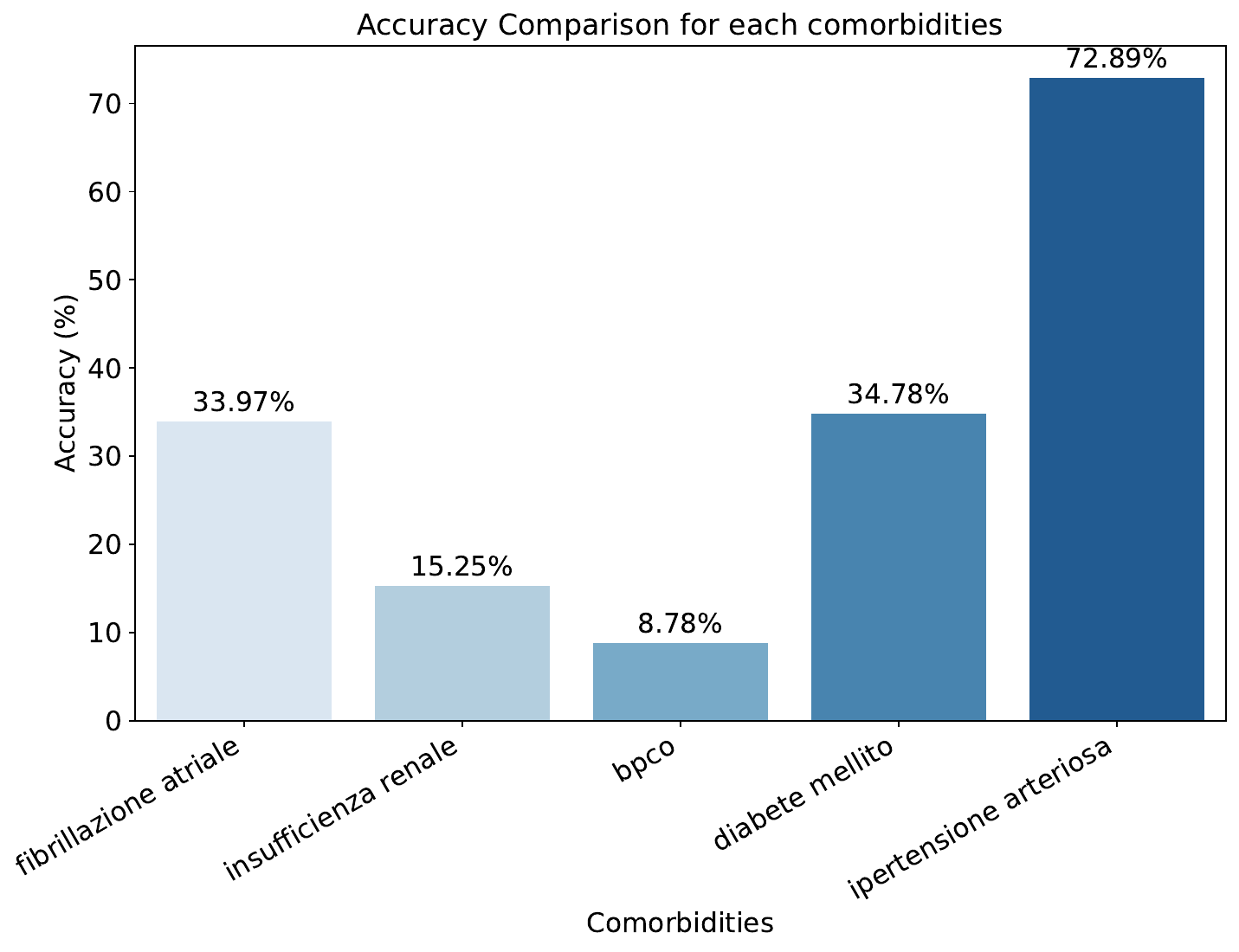}  
        \caption{}
    \end{subfigure}
    \begin{subfigure}[b]{0.3\textwidth}
        \includegraphics[width=\linewidth]{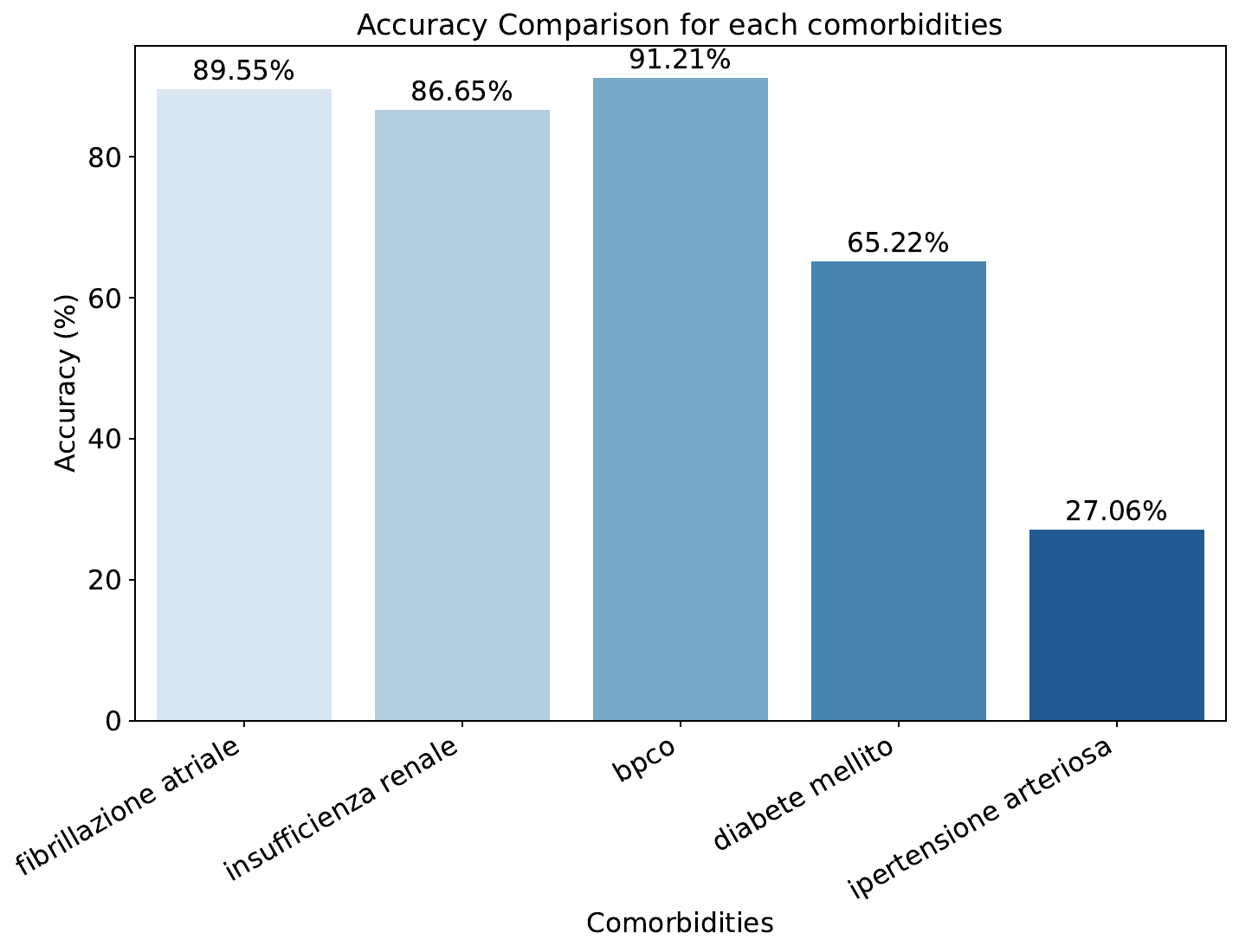}  
        \caption{}
    \end{subfigure}
    \begin{subfigure}[b]{0.3\textwidth}
        \includegraphics[width=\linewidth]{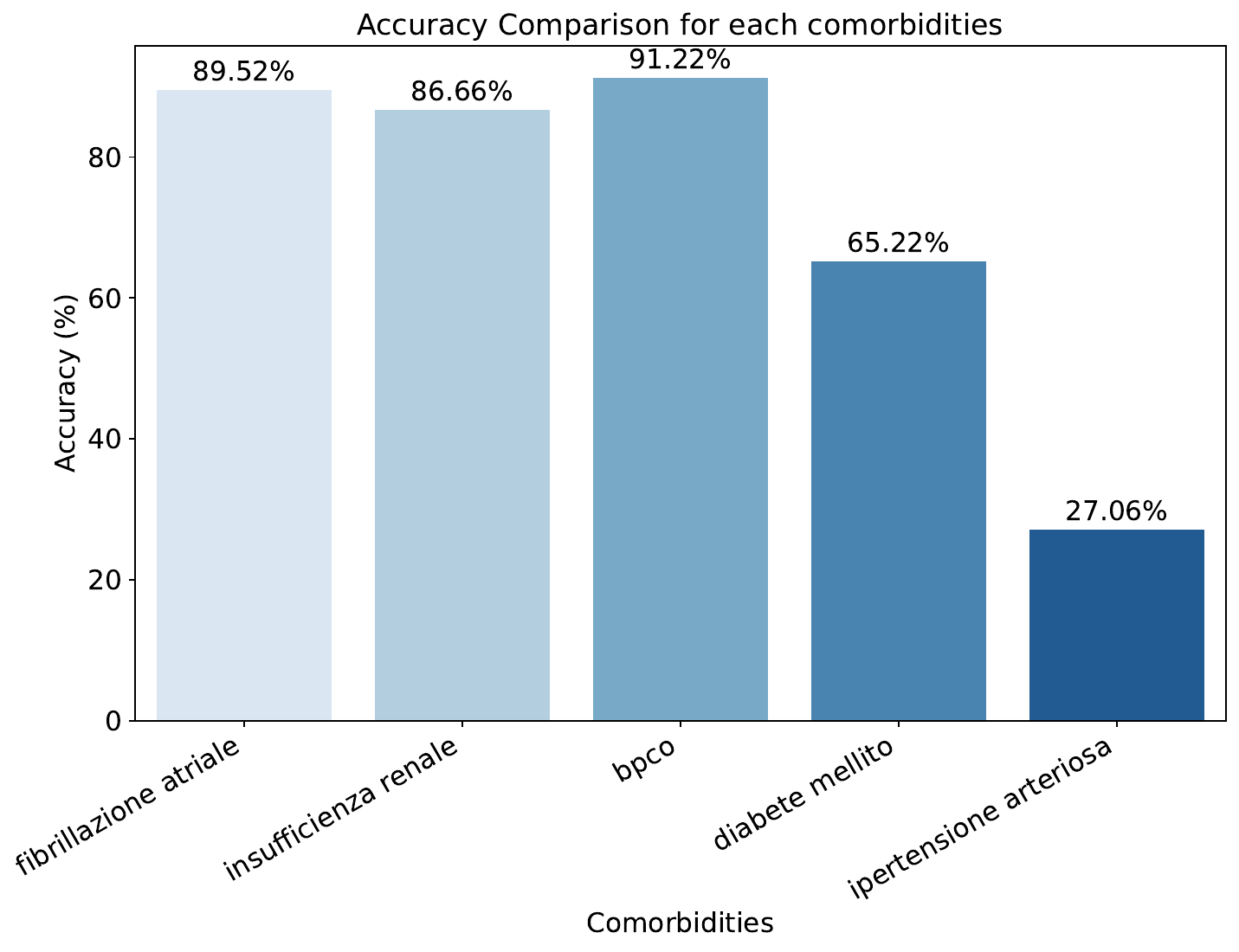}  
        \caption{}
    \end{subfigure}
    \caption{LLMs accuracy compared to regular expression -a)OpenLLaMA 3B, b)OpenLLaMA 7B, c)Mistral 7B, d)Mixtral 8x7B, e)Qwen2.5 3B and e)Qwen2.5 7B}\label{fig:regexp1}
\end{figure}

$OpenLLaMA$ model family generally show a increase in the classification accuracy w.r.t regular expression results of the comorbidities as the model size increases, which can been seen from the Figure~\ref{fig:regexp1}. In particular with OpenLLaMA 3B, the model tends to perform well in terms of comorbidities like Diabete mellito and Ipertensione arteriosa with an accuracy of 34.78~\% and 72.86~\% compared to the other comorbidities which have an classification accuracy of less than 15~\%. In contrast OpenLLaMA 7B has an accuracy of 50~\% and above across all the comorbidities, Insufficienza Renale and BPCO have been among the top in classification with 83.73~\% and 81.95~\%. $Mistral$ model family generally show a decrease in the classification accuracy w.r.t regular expression results of the comorbidities as the model size increases, which is shown in Figure~\ref{fig:regexp1}. Mistral 7B shows a accuracy of 75~\% above across all the comorbidities except Ipertensione arteriosa which is at 57.33~\%, Fibrillazione atriale and BPCO have shown a accucary of 90~\% above which is the highest among all the models. Mixtral 8x7B shows a varied performance, comorbidities like BPCO, Diabete mellito and Ipertensione arteriosa classification metrics are more equivalent to the performance OpenLLaMA 3B. Fibrillazione atriale and Insufficienza Renale show a slight increase in accuracy when compared to OpenLLaMA 3B. In contrast $Qwen2.5$ model family does not show much of a increase or decrease with increase in the model parameters, which is shown in the Figure~\ref{fig:regexp1}. Qwen2.5 7B which shows a very small marginal difference w.r.t to the 3B model, especially only for comorbidities like Fibrillazione atriale, Insufficienza Renale and BPCO. This shows that the domain-specific features in the dataset are being handled similarly by both models and the differences in model size is not substantial enough to affect their performance.

\begin{figure}[h]
    \centering
    \begin{subfigure}[b]{0.4\textwidth}
        \includegraphics[width=\linewidth,trim=30pt 30pt 30pt 30pt, clip]{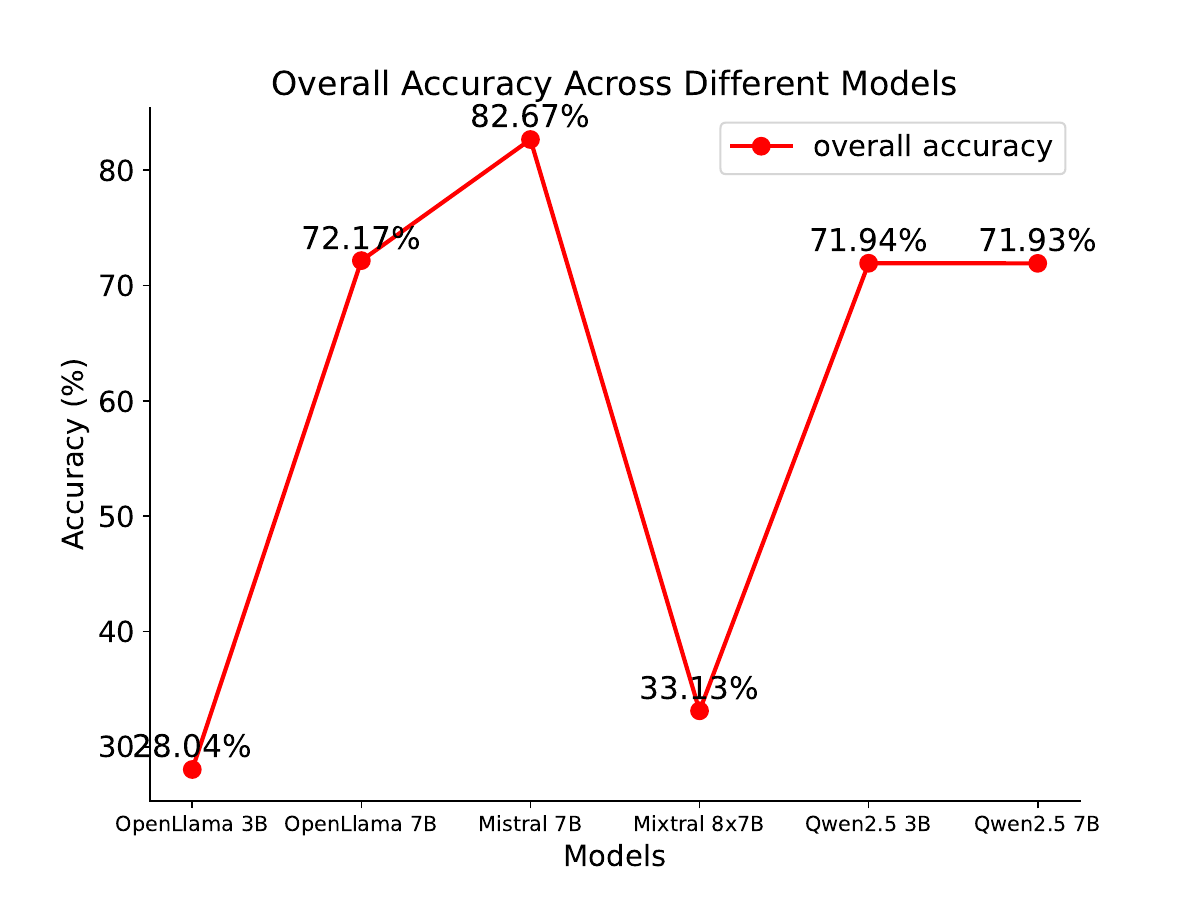}  
        \caption{}
    \end{subfigure}
    \begin{subfigure}[b]{0.4\textwidth}
        \includegraphics[width=\linewidth,trim=30pt 30pt 30pt 30pt, clip]{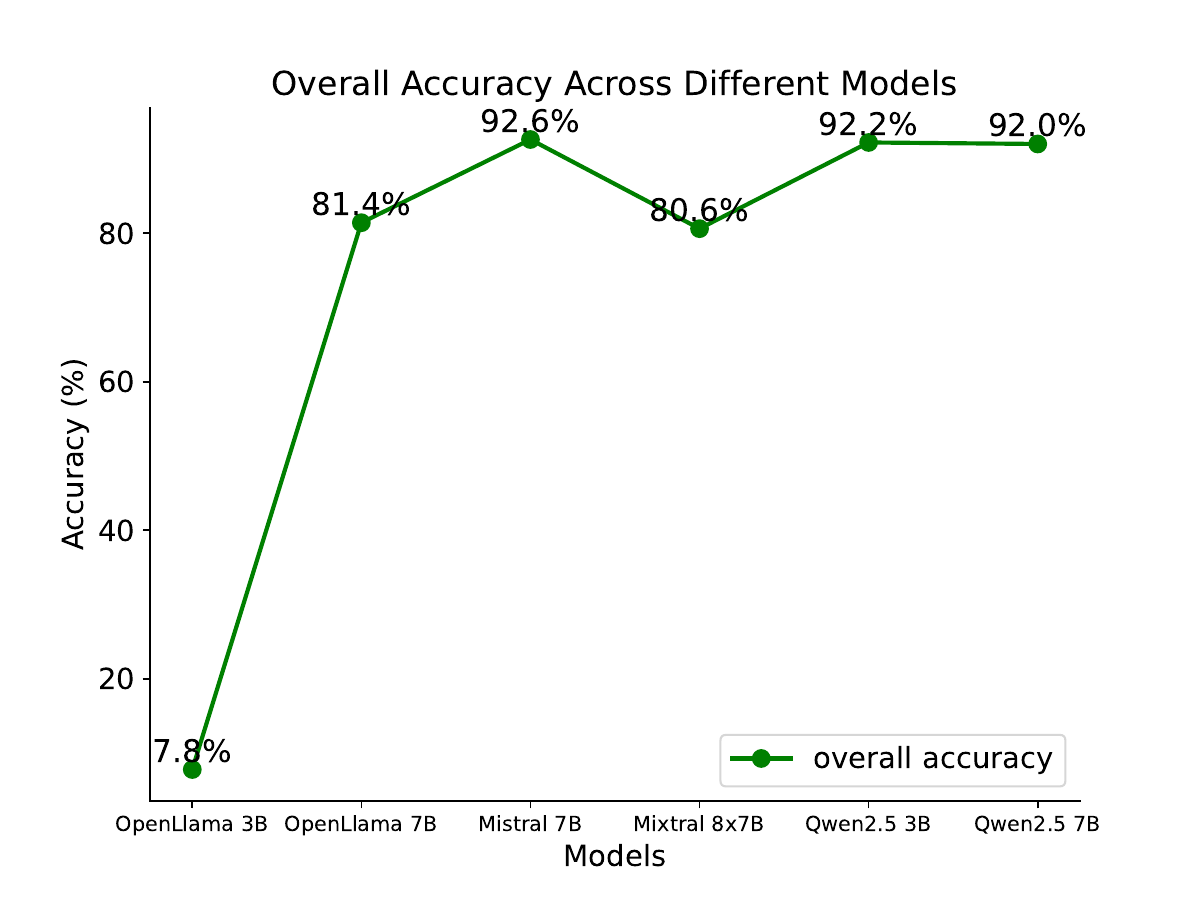}  
        \caption{}
    \end{subfigure}
    \caption{Overall accuracy across different models compared to a)regular expression annotation and b) manual annotation.}\label{fig:llm_both}
\end{figure}

From the Figure~\ref{fig:llm_both}, we could compare the overall accuracy of the selected models to the a) automated and b) manual annotated data as the reference values. It shows the mean accuracy of the models across all the defined comorbidities. From a) of the Figure~\ref{fig:llm_both} compared to the automated annotation, we could see that OpenLlama 3B and Mixtral 8x7B overall accuracies were lower than 35~\%. In contrast, OpenLlama 7B, Mistral 7B, Qwen2.5 3B and 7B both had an overall accuracy of more than 70~\%, which is nearly two times the performance of the other two models. Mistral 7B performed very well when compared to the other models, with an overall accuracy of 82.67~\% when compared to regular expression (automated annotation). 

While the classification accuracy provides a general insight into the LLMs performance, it is important to examine the classification reports for a deeper understanding. Metrics such as F1-score, precision and recall help to assess how well the model performs across different comorbidities especially in the presence of data imbalance, and highlight potential areas where it may be struggling. This is needed for evaluating the LLMs true effectiveness.

\begin{table}[h]
\resizebox{\columnwidth}{!}{
\begin{tabular}{|cc|ccccc|}
\hline
\multicolumn{2}{|c|}{\multirow{2}{*}{\textbf{LLMs and Metrics}}}                  & \multicolumn{5}{c|}{\textbf{Automated (8223 data records)}}                                                                                                                                                                     \\ \cline{3-7} 
\multicolumn{2}{|c|}{}                                                            & \multicolumn{1}{c|}{\textbf{Fibrillazione Atriale}} & \multicolumn{1}{c|}{\textbf{Insufficienza Renale}} & \multicolumn{1}{c|}{\textbf{BPCO}} & \multicolumn{1}{c|}{\textbf{Diabete Mellito}} & \textbf{Ipertensione Arteriosa} \\ \hline
\multicolumn{1}{|c|}{\multirow{3}{*}{\textbf{OpenLLaMA 3B}}} & \textbf{Precision} & \multicolumn{1}{c|}{0.1}                            & \multicolumn{1}{c|}{0.13}                          & \multicolumn{1}{c|}{0.09}          & \multicolumn{1}{c|}{0.35}                     & 0.73                            \\ \cline{2-7} 
\multicolumn{1}{|c|}{}                                       & \textbf{Recall}    & \multicolumn{1}{c|}{1}                              & \multicolumn{1}{c|}{1}                             & \multicolumn{1}{c|}{1}             & \multicolumn{1}{c|}{1}                        & 1                               \\ \cline{2-7} 
\multicolumn{1}{|c|}{}                                       & \textbf{F1-score}  & \multicolumn{1}{c|}{0.19}                           & \multicolumn{1}{c|}{0.24}                          & \multicolumn{1}{c|}{0.16}          & \multicolumn{1}{c|}{0.52}                     & 0.84                            \\ \hline
\multicolumn{1}{|c|}{\multirow{3}{*}{\textbf{OpenLLaMA 7B}}} & \textbf{Precision} & \multicolumn{1}{c|}{0.23}                           & \multicolumn{1}{c|}{0.19}                          & \multicolumn{1}{c|}{0.22}          & \multicolumn{1}{c|}{0.46}                     & 0.8                             \\ \cline{2-7} 
\multicolumn{1}{|c|}{}                                       & \textbf{Recall}    & \multicolumn{1}{c|}{0.52}                           & \multicolumn{1}{c|}{0.07}                          & \multicolumn{1}{c|}{0.42}          & \multicolumn{1}{c|}{0.47}                     & 0.53                            \\ \cline{2-7} 
\multicolumn{1}{|c|}{}                                       & \textbf{F1-score}  & \multicolumn{1}{c|}{0.31}                           & \multicolumn{1}{c|}{0.1}                           & \multicolumn{1}{c|}{0.29}          & \multicolumn{1}{c|}{0.47}                     & 0.64                            \\ \hline
\multicolumn{1}{|c|}{\multirow{3}{*}{\textbf{Mistral 7B}}}   & \textbf{Precision} & \multicolumn{1}{c|}{0.86}                           & \multicolumn{1}{c|}{0.96}                          & \multicolumn{1}{c|}{1}             & \multicolumn{1}{c|}{0.99}                     & 0.99                            \\ \cline{2-7} 
\multicolumn{1}{|c|}{}                                       & \textbf{Recall}    & \multicolumn{1}{c|}{0.69}                           & \multicolumn{1}{c|}{0.11}                          & \multicolumn{1}{c|}{0.15}          & \multicolumn{1}{c|}{0.42}                     & 0.42                            \\ \cline{2-7} 
\multicolumn{1}{|c|}{}                                       & \textbf{F1-score}  & \multicolumn{1}{c|}{0.77}                           & \multicolumn{1}{c|}{0.2}                           & \multicolumn{1}{c|}{0.26}          & \multicolumn{1}{c|}{0.59}                     & 0.59                            \\ \hline
\multicolumn{1}{|c|}{\multirow{3}{*}{\textbf{Mixtral 8x7B}}} & \textbf{Precision} & \multicolumn{1}{c|}{0.10}                           & \multicolumn{1}{c|}{0.13}                          & \multicolumn{1}{c|}{0.09}          & \multicolumn{1}{c|}{0.35}                     & 0.73                            \\ \cline{2-7} 
\multicolumn{1}{|c|}{}                                       & \textbf{Recall}    & \multicolumn{1}{c|}{0.69}                           & \multicolumn{1}{c|}{0.98}                          & \multicolumn{1}{c|}{1}             & \multicolumn{1}{c|}{1}                        & 1                               \\ \cline{2-7} 
\multicolumn{1}{|c|}{}                                       & \textbf{F1-score}  & \multicolumn{1}{c|}{0.18}                           & \multicolumn{1}{c|}{0.24}                          & \multicolumn{1}{c|}{0.16}          & \multicolumn{1}{c|}{0.52}                     & 0.84                            \\ \hline
\multicolumn{1}{|c|}{\multirow{3}{*}{\textbf{Qwen2.5 3B}}}   & \textbf{Precision} & \multicolumn{1}{c|}{0}                              & \multicolumn{1}{c|}{0}                             & \multicolumn{1}{c|}{0}             & \multicolumn{1}{c|}{0}                        & 0                               \\ \cline{2-7} 
\multicolumn{1}{|c|}{}                                       & \textbf{Recall}    & \multicolumn{1}{c|}{0}                              & \multicolumn{1}{c|}{0}                             & \multicolumn{1}{c|}{0}             & \multicolumn{1}{c|}{0}                        & 0                               \\ \cline{2-7} 
\multicolumn{1}{|c|}{}                                       & \textbf{F1-score}  & \multicolumn{1}{c|}{0}                              & \multicolumn{1}{c|}{0}                             & \multicolumn{1}{c|}{0}             & \multicolumn{1}{c|}{0}                        & 0                               \\ \hline
\multicolumn{1}{|c|}{\multirow{3}{*}{\textbf{Qwen2.5 7B}}}   & \textbf{Precision} & \multicolumn{1}{c|}{0.25}                           & \multicolumn{1}{c|}{0}                             & \multicolumn{1}{c|}{0}             & \multicolumn{1}{c|}{0}                        & 0                               \\ \cline{2-7} 
\multicolumn{1}{|c|}{}                                       & \textbf{Recall}    & \multicolumn{1}{c|}{0}                              & \multicolumn{1}{c|}{0}                             & \multicolumn{1}{c|}{0}             & \multicolumn{1}{c|}{0}                        & 0                               \\ \cline{2-7} 
\multicolumn{1}{|c|}{}                                       & \textbf{F1-score}  & \multicolumn{1}{c|}{0.1}                            & \multicolumn{1}{c|}{0}                             & \multicolumn{1}{c|}{0}             & \multicolumn{1}{c|}{0}                        & 0                               \\ \hline
\end{tabular}
}
\caption{Precision, Recall and F1 Score for LLMs vs. Regular Expressions in Identifying Comorbidities (Class 1).}\label{table:class_report}
\end{table}

From the Table~\ref{table:class_report} we could compare comorbidity-wise LLMs performance. OpenLLama 3B has perfect recall (1.0) but struggles with precision, leading to many false positives. OpenLLaMA 7B model shows low precision in most of the conditions i.e, ranging from 0.19 for Insufficienza Renale to 0.8 for Ipertensione Arteriosa. OpenLLama 3B with high recall, low precision and low F1, the model struggles to generalization across comorbidities. OpenLLama 7B shows better generalization than 3B, but still has low scores across various comorbidities. 

Mistral 7B demonstrates outstanding precision across all comorbidities, which shows it is very effective at avoiding false positives. But recall is low compared to high precision, indicating the model often misses many cases of the comorbidities. In terms of F1-score, the model shows better balanced results across all comorbidities, compared to other LLMs.
Mixtral 8x7B model exhibits low precision and high recall across all comorbidities. Due to this, the F1-scores are low for all except for Ipertensione Arteriosa is 0.84, showing the model achieves a better balance between precision and recall for this comorbidity.
Mistral 7B shows better generalization, but still has low scores across a few comorbidities. Mixtral 8x7B has low scores in most cases and struggles to generalize across comorbidities, even though it is the most advanced among all the chosen models.

Both Qwen2.5 3B and Qwen2.5 7B models, despite their parameter differences, seem to produce classes that do not match the true positive class at all in most of the cases. The precision is 0 in most of the cases because there are no positive predictions. Similarly, recall is also 0 because the model does not correctly identify any of the actual positive samples in the data. In context with these, F1-score is also 0 for most of the comorbidities. Qwen2.5 7B exhibits a slight difference for Fibrillazione Atriale with precision(0.25) and F1-score(0.1). Overall, Qwen2.5 family could not produce true positive class when automated data was compared.

\begin{tcolorbox}
[width=\linewidth, sharp corners=all, colback=white!95!black]
 LLMs in a zero-shot setting struggle to extract comorbidities from EHRs and do not match the accuracy of regular expression-based extractions, making them unsuitable as a substitute.
 These results are linked to research questions \textbf{Q1)} and \textbf{Q2)}. 
\end{tcolorbox}

\subsection{Performance Comparison: LLMs vs Humans}
As discussed in section~\ref{methodology} 100 $false$ classified comorbidities has been manually annotated by clinical experts. This is because positive classifications contain comorbidity information in the EHRs. False negatives in regexp are critical in healthcare, as missed diagnoses can delay treatment, progression of disease and lead to poor patient outcomes. By annotating these false negatives, we can ensure that critical misclassifications are corrected, improving the ability to identify important conditions. 

\begin{figure}[h]
    \centering
    \begin{subfigure}[b]{0.19\textwidth}
        \includegraphics[width=\linewidth]{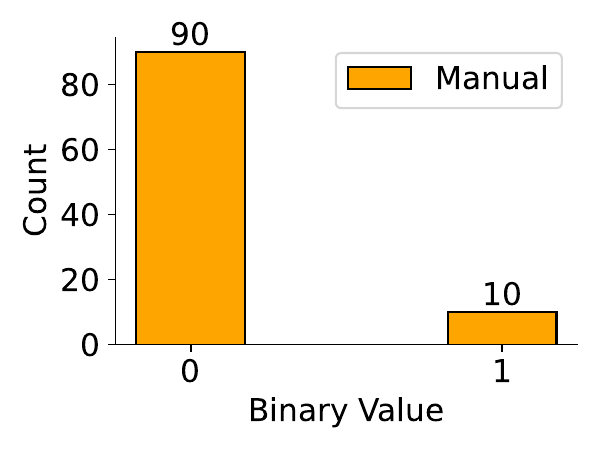}  
        \caption{}
    \end{subfigure}
    \begin{subfigure}[b]{0.19\textwidth}
        \includegraphics[width=\linewidth]{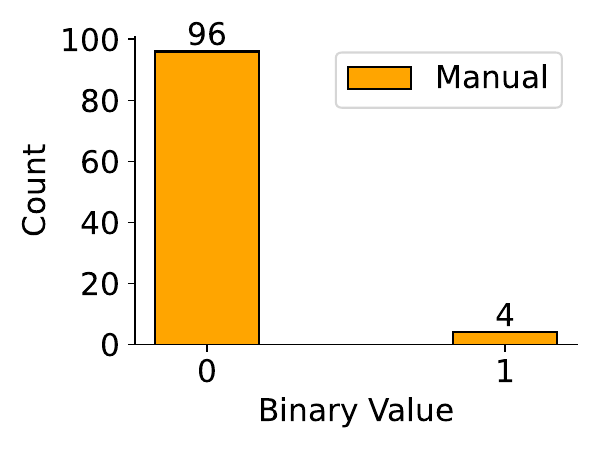}  
        \caption{}
    \end{subfigure}
    \begin{subfigure}[b]{0.19\textwidth}
        \includegraphics[width=\linewidth]{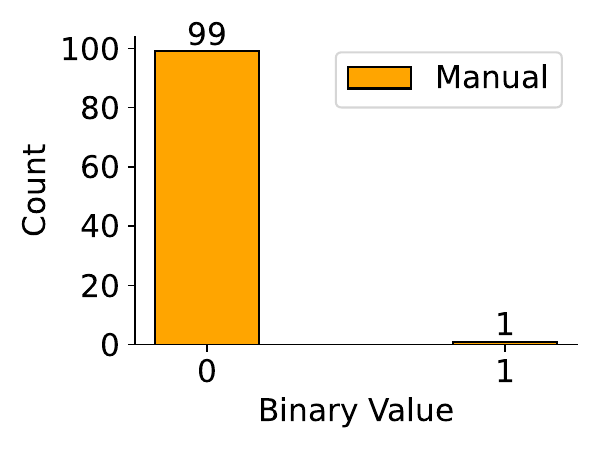}  
        \caption{}
    \end{subfigure}
    \begin{subfigure}[b]{0.19\textwidth}
        \includegraphics[width=\linewidth]{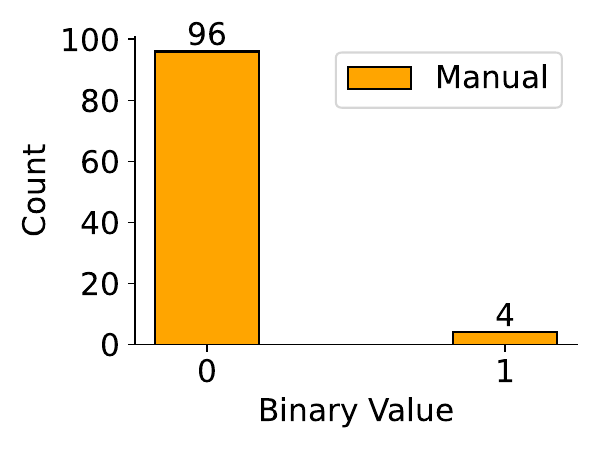}  
        \caption{}
    \end{subfigure}
    \begin{subfigure}[b]{0.19\textwidth}
        \includegraphics[width=\linewidth]{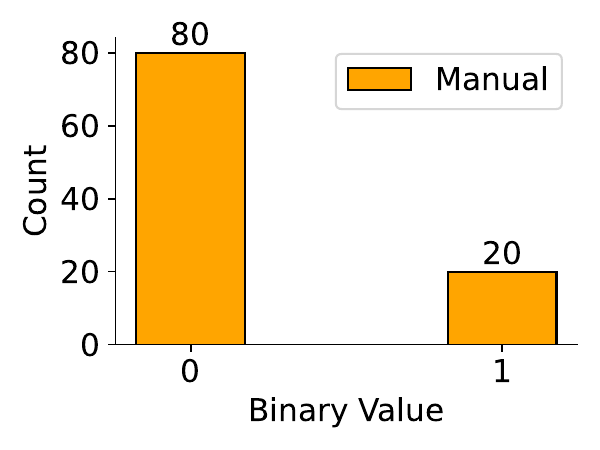}  
        \caption{}
    \end{subfigure}
    \caption{Manual annotation classification of the chosen comorbidities - a)Fibrillazione atriale, b)Insufficienza Renale, c)BPCO-Broncopneumopatia cronica ostruttiva, d)Diabete mellito and e)Ipertensione arteriosa.}\label{fig:man_anno}
\end{figure}

During this process, clinicians were not informed about the nature of the dataset provided i.e., with only the false class from the regex annotations provided. This was done to avoid bias or prejudgment from influencing the annotation. The guidance provided to the annotators is same as before, 0 class to represent the comorbidities is not found and
1 class represent the availability the comorbidities in the EHRs. From the Figure~\ref{fig:man_anno} we could see that Ipertensione arteriosa is the one with more false negatives, followed by the Fibrillazione atriale. Both of these comorbidities have 10~\% or more false classification by the regular expression. In the other hand Insufficienza Renale, BPCO and Diabete mellito have 4~\% or less false negative compared to the counterpart. In addition to this we can derive the performance of the regular expression created for each comorbidities when it is compared to the manual annotation, in particular pattern created for BPCO has the higher accuracy of 99~\% and 80~\% for Ipertensione arteriosa being the least accuracy. Overall comorbidities classification accuracy using regular expression when compared to the manual annotation was 92.2~\%.  

\textbf{LLMs}. With this new set of data, a zero-shot setting with a standard prompt is used across all the selected LLMs. As per previous experiment, to prevent multiple classifications in a single inference, each comorbidity is classified individually for each EHR. In the following we discusses the classification results of various LLM families against manual annotations.

\begin{figure}[h]
    \centering
    % \resizebox{0.9\textwidth}{!}{
    \begin{subfigure}[b]{0.3\textwidth}
        \includegraphics[width=\linewidth]{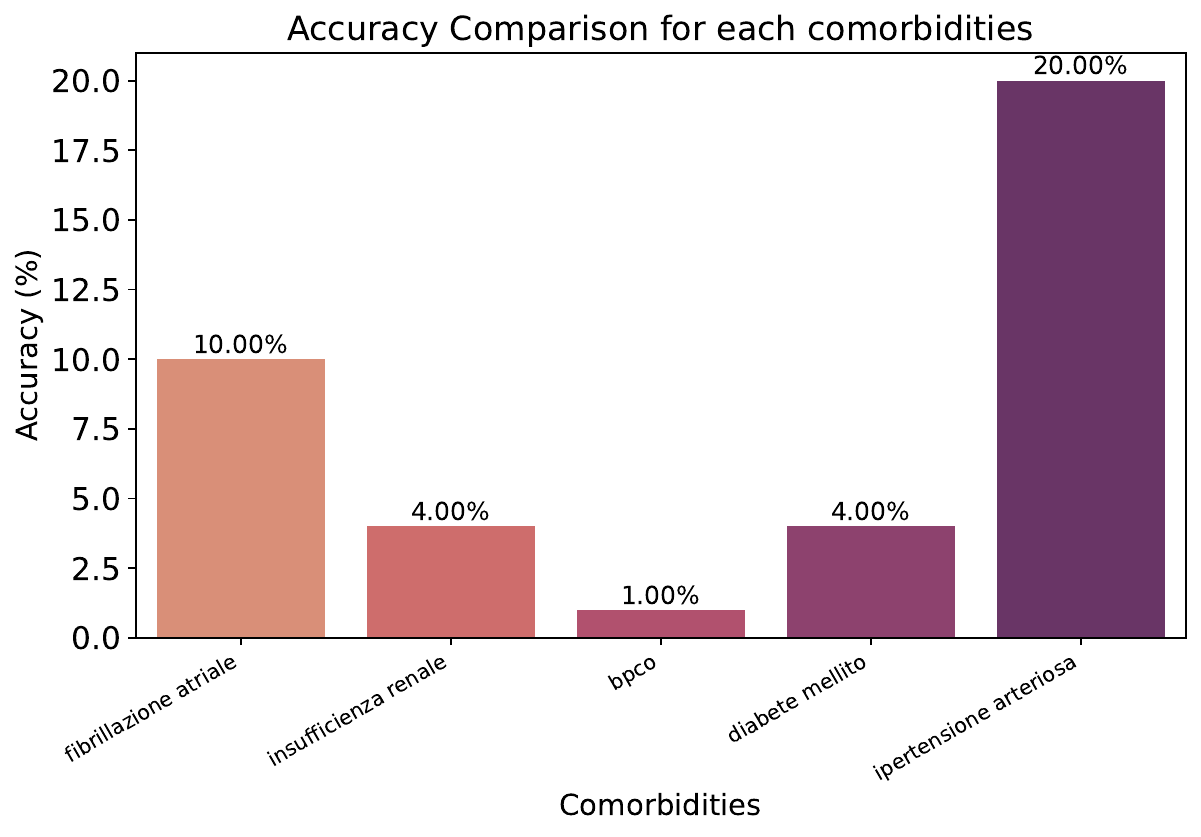}  
        \caption{}
    \end{subfigure}
    \begin{subfigure}[b]{0.3\textwidth}
        \includegraphics[width=\linewidth]{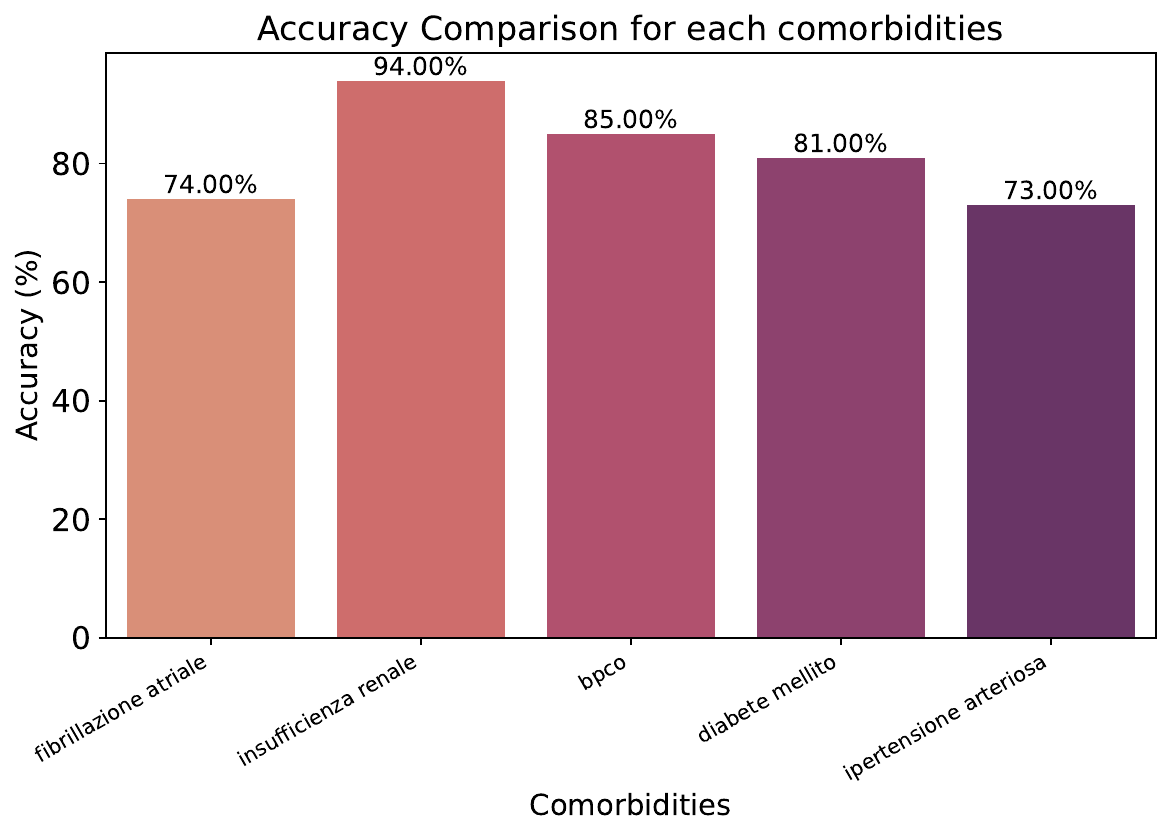}  
        \caption{}
    \end{subfigure}
    \begin{subfigure}[b]{0.3\textwidth}
        \includegraphics[width=\linewidth]{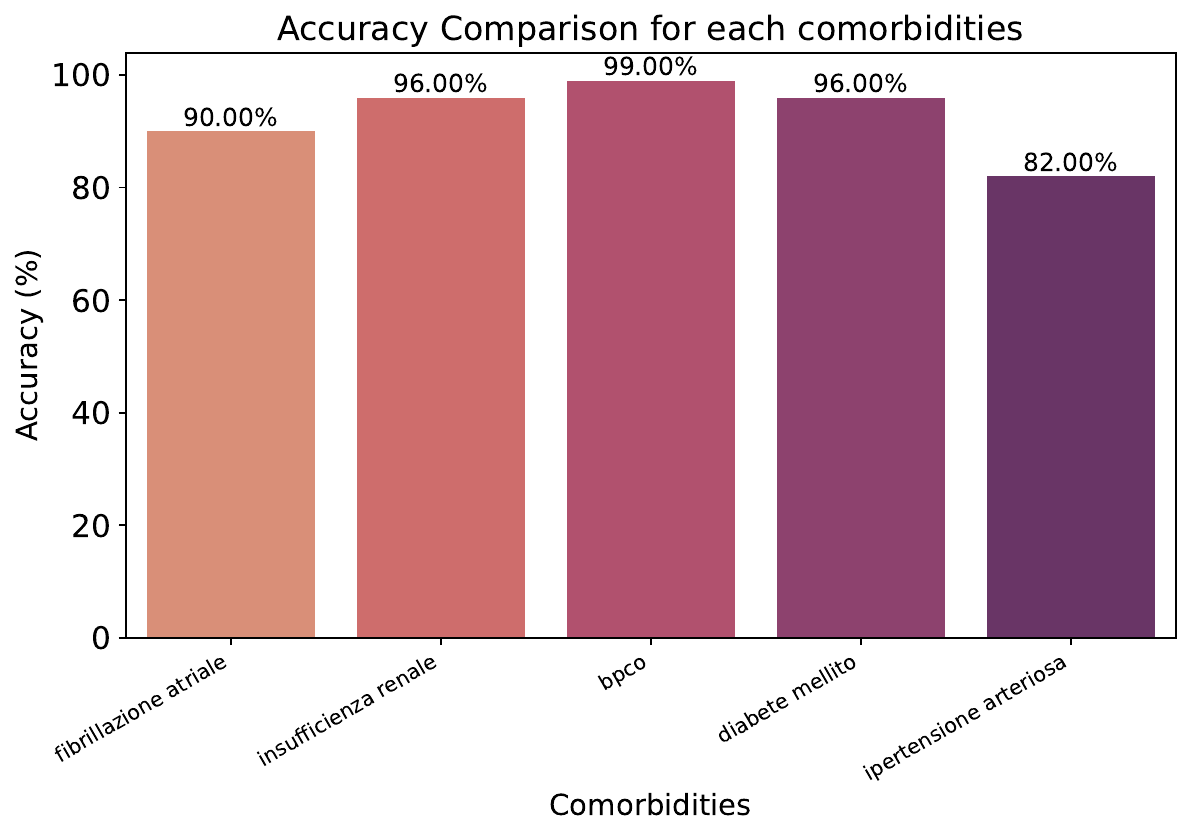}  
        \caption{}
    \end{subfigure}
    \begin{subfigure}[b]{0.3\textwidth}
        \includegraphics[width=\linewidth]{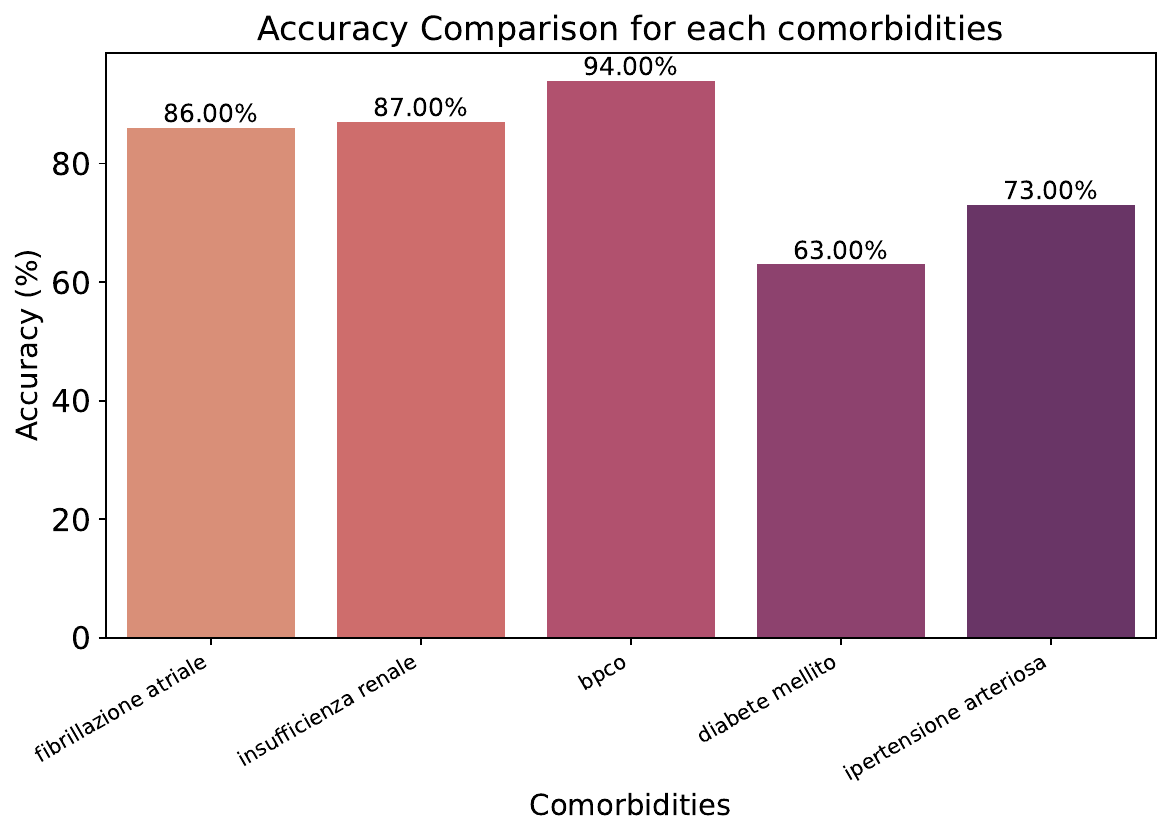}  
        \caption{}
    \end{subfigure}
    \begin{subfigure}[b]{0.3\textwidth}
        \includegraphics[width=\linewidth]{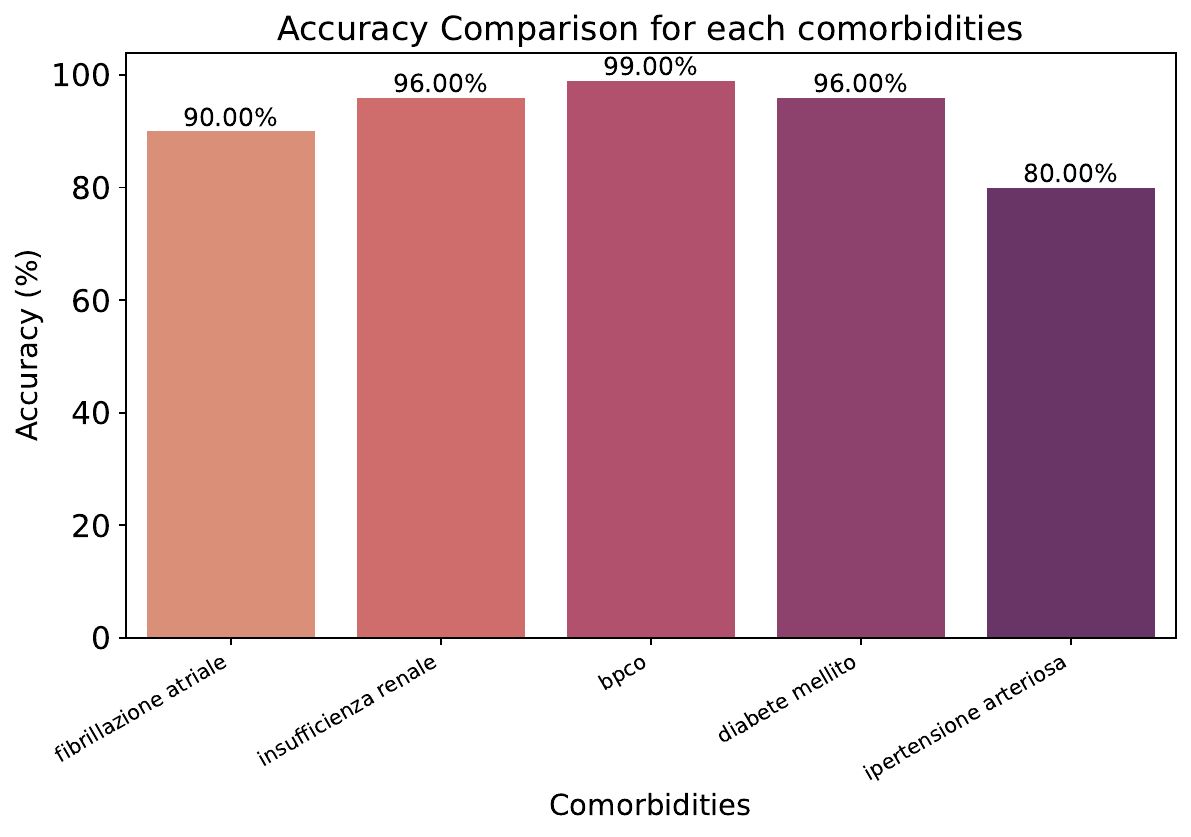}  
        \caption{}
    \end{subfigure}
    \begin{subfigure}[b]{0.3\textwidth}
        \includegraphics[width=\linewidth]{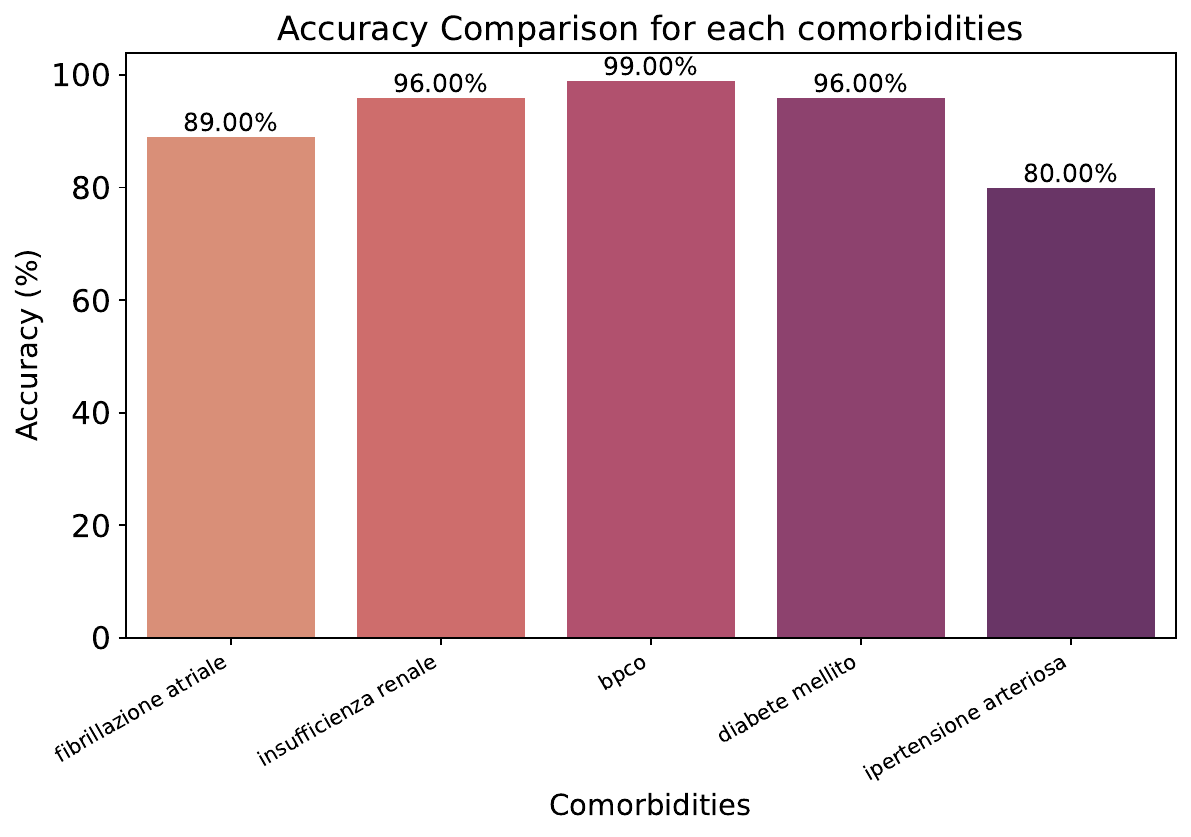}  
        \caption{}
    \end{subfigure}
    % }
    \caption{LLMs accuracy compared to manual annotation. (a)OpenLLaMA 3B, (b)OpenLLaMA 7B, (c)Mistral 7B, (d)Mixtral 8x7B, (e)Qwen2.5 3B and (f)Qwen2.5 7B}\label{fig:regexp2}
\end{figure}

$OpenLLaMA$ family of models shows nearly the same kind of behavior as seen with the regular expression, increase in the model accuracy w.r.t the model size which can be seen in the Figure~\ref{fig:regexp2}. OpenLLaMA 3B has varying range of classification accuracy when compared to the manual annotation set, starting from BPCO as low as 1~\% to 20~\% for Ipertensione arteriosa. In contrast OpenLLaMA 7B shows a extreme level of accuracy increase compared to the 3B version. In particular the model shows classification accuracy of greater than 70~\% across all the comorbidities,  notably 94~\% accuracy for Insufficienza Renale and 85~\% for BPCO. These results show that the model are mostly in alignment with the human annotation. $Mistral$ family shows increased level of accuracy w.r.t to the previous regular expression case, the model Mixtral 8x7B showed a drastic performance increase in manual annotation. Mistral 7B shows 90~\% and above classification accuracy in most of the comorbidities except  for Ipertensione arteriosa which is 82~\% and reaching 99~\% for BPCO. Mistral 8x7B has a 94~\% for BPCO and 63~\% for Diabete mellito being the least, the model shows a overall performance increase across all the comorbidities when compared to the manual annotation. $Qwen2.5$ family shows increased level of accuracy  w.r.t to the regular expression case, notably a huge increase 52.94\% for Ipertensione arteriosa and 30.78\% for Diabete mellito. Qwen2.5 3B and 7B differ slightly in the Fibrillazione atriale and other comorbidities classification accuracy being identical.

\begin{figure}[t]
    \centering

    \begin{subfigure}[b]{\textwidth}
    \setcounter{subfigure}{0}
        \centering
    \begin{subfigure}[b]{0.19\textwidth}
        \includegraphics[width=\linewidth]{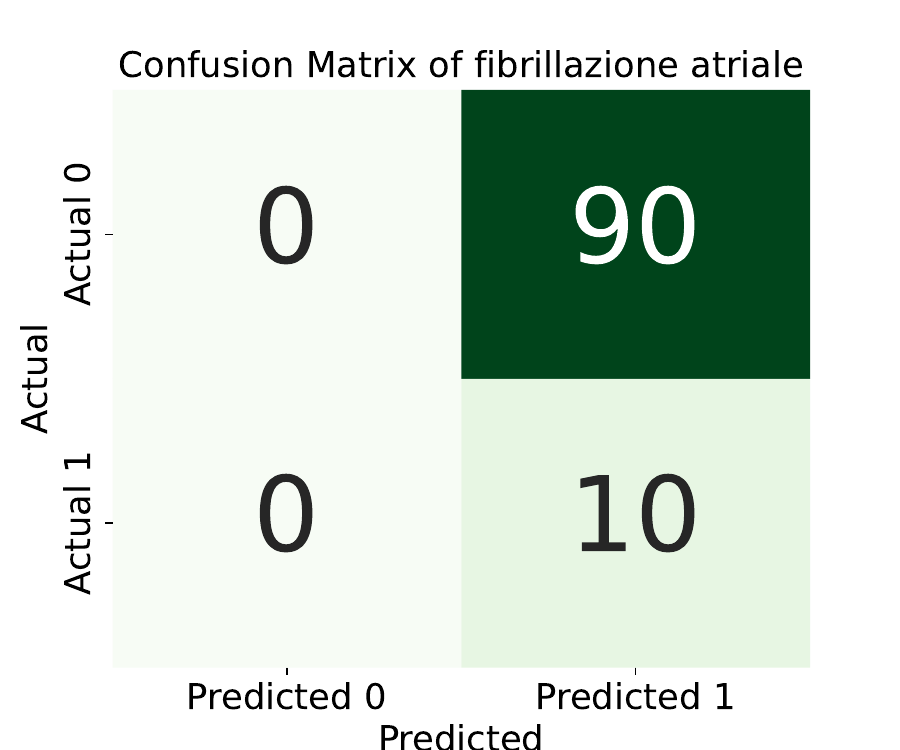} 
        \caption{}
    \end{subfigure}
    \begin{subfigure}[b]{0.19\textwidth}
        \includegraphics[width=\linewidth]{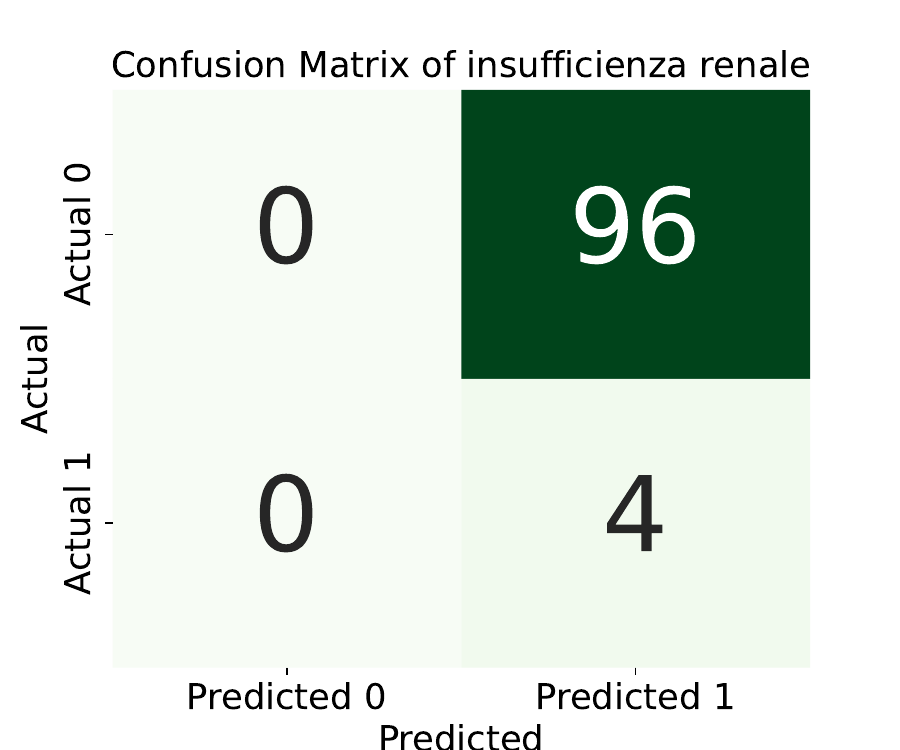}  
        \caption{}
    \end{subfigure}
    \begin{subfigure}[b]{0.19\textwidth}
        \includegraphics[width=\linewidth]{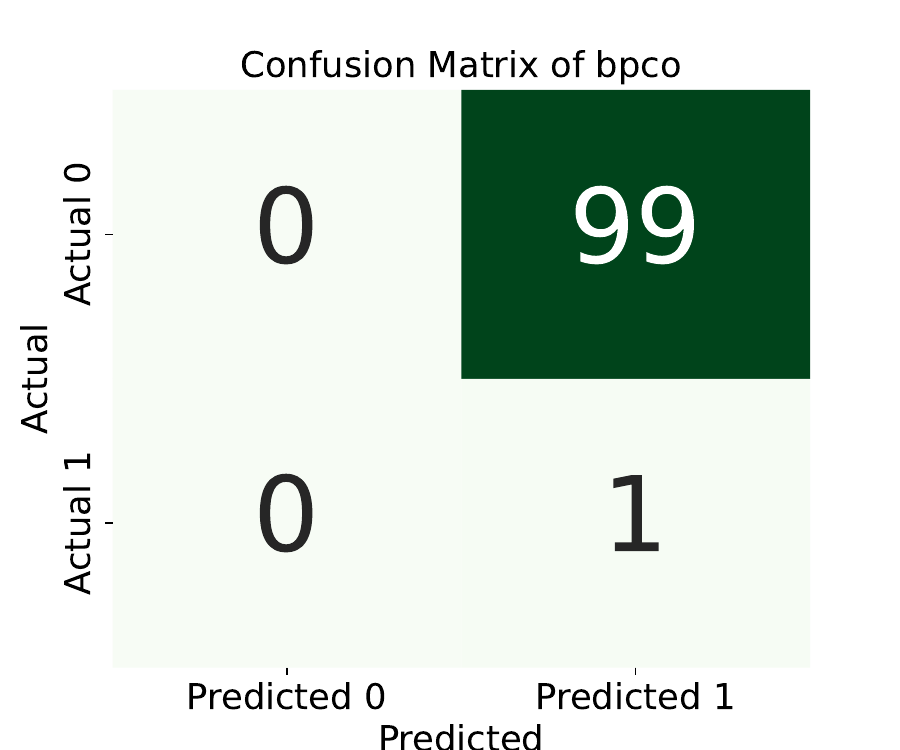}  
        \caption{}
    \end{subfigure}
    \begin{subfigure}[b]{0.19\textwidth}
        \includegraphics[width=\linewidth]{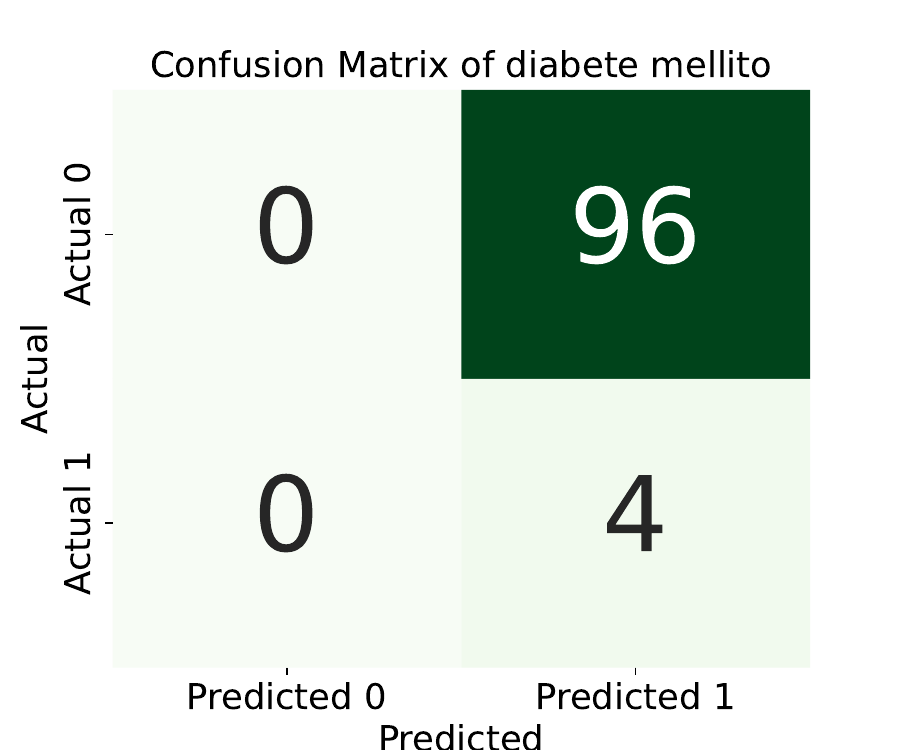}  
        \caption{}
    \end{subfigure}
    \begin{subfigure}[b]{0.19\textwidth}
        \includegraphics[width=\linewidth]{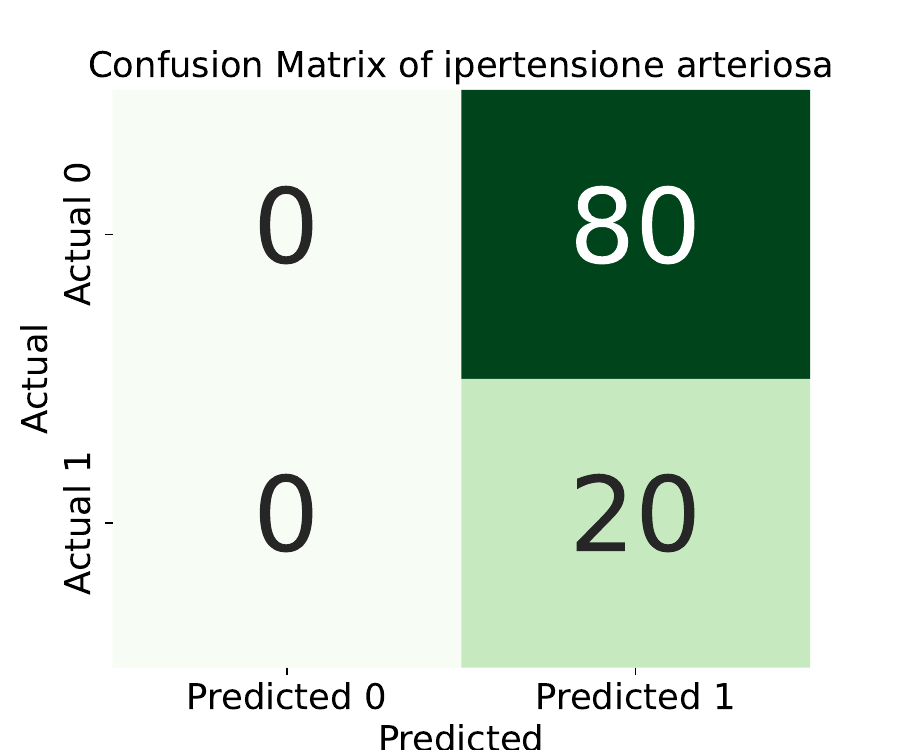}  
        \caption{}
    \end{subfigure}  
    \begin{subfigure}[b]{0.19\textwidth}
        \includegraphics[width=\linewidth]{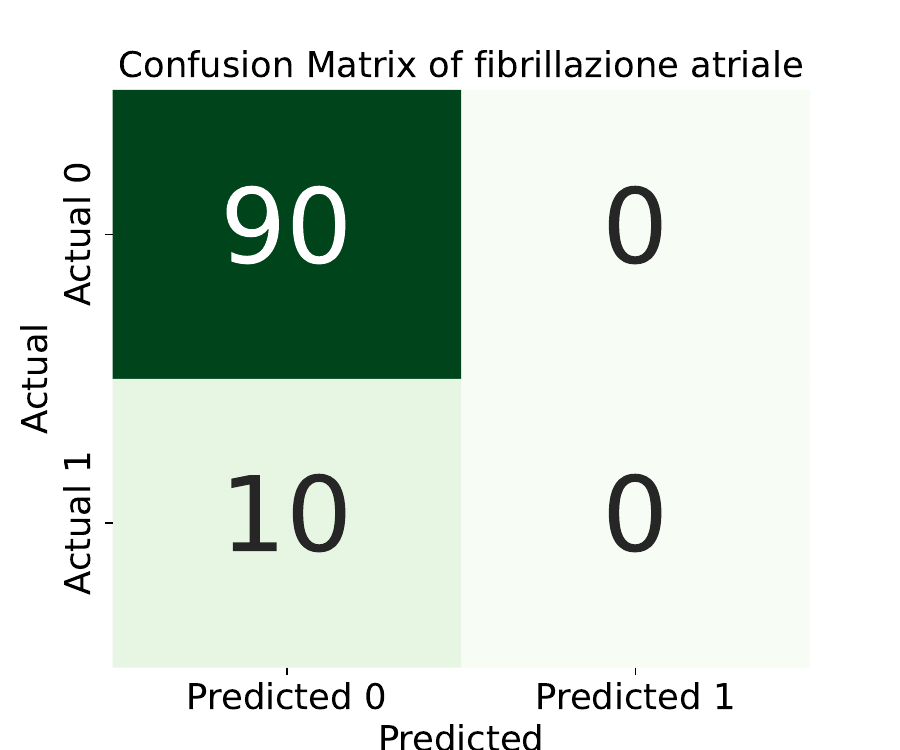}  
        \caption{}
    \end{subfigure}
    \begin{subfigure}[b]{0.19\textwidth}
        \includegraphics[width=\linewidth]{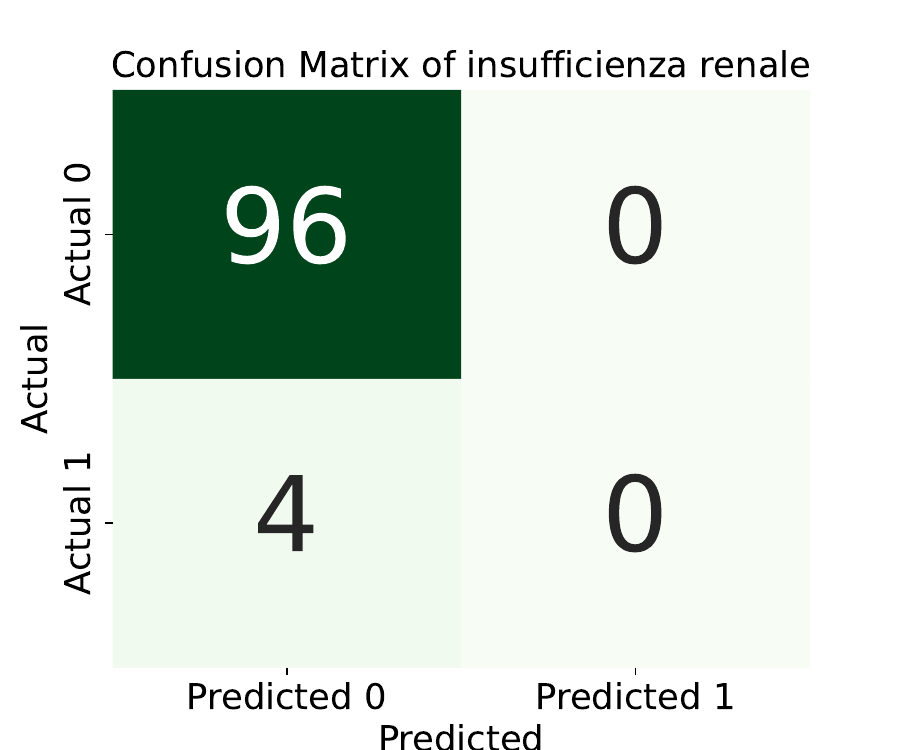}  
        \caption{}
    \end{subfigure}
    \begin{subfigure}[b]{0.19\textwidth}
        \includegraphics[width=\linewidth]{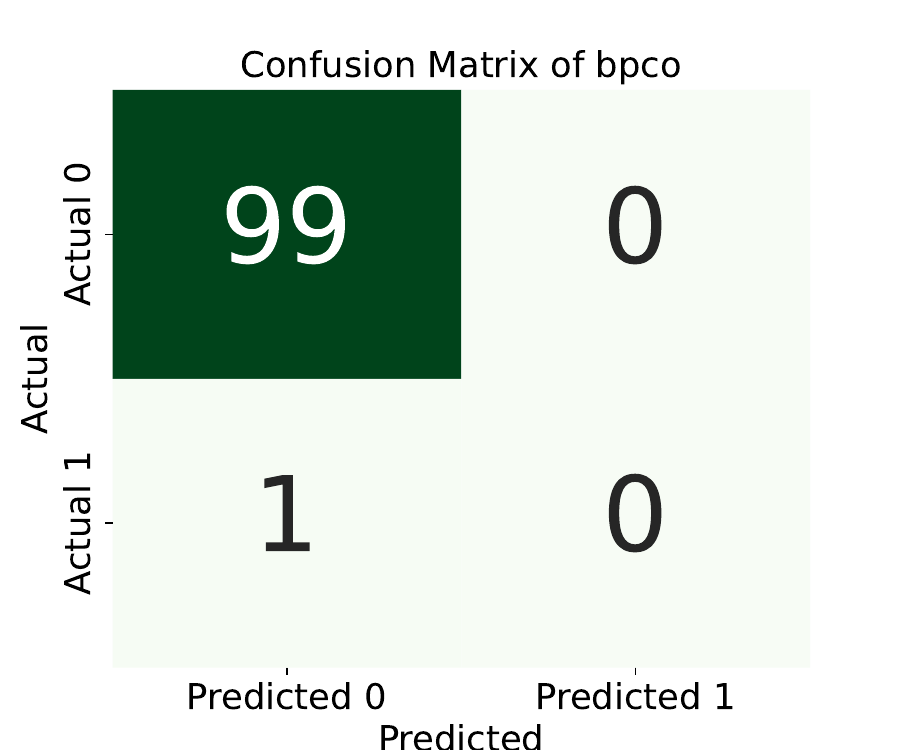}  
        \caption{}
    \end{subfigure}
    \begin{subfigure}[b]{0.19\textwidth}
        \includegraphics[width=\linewidth]{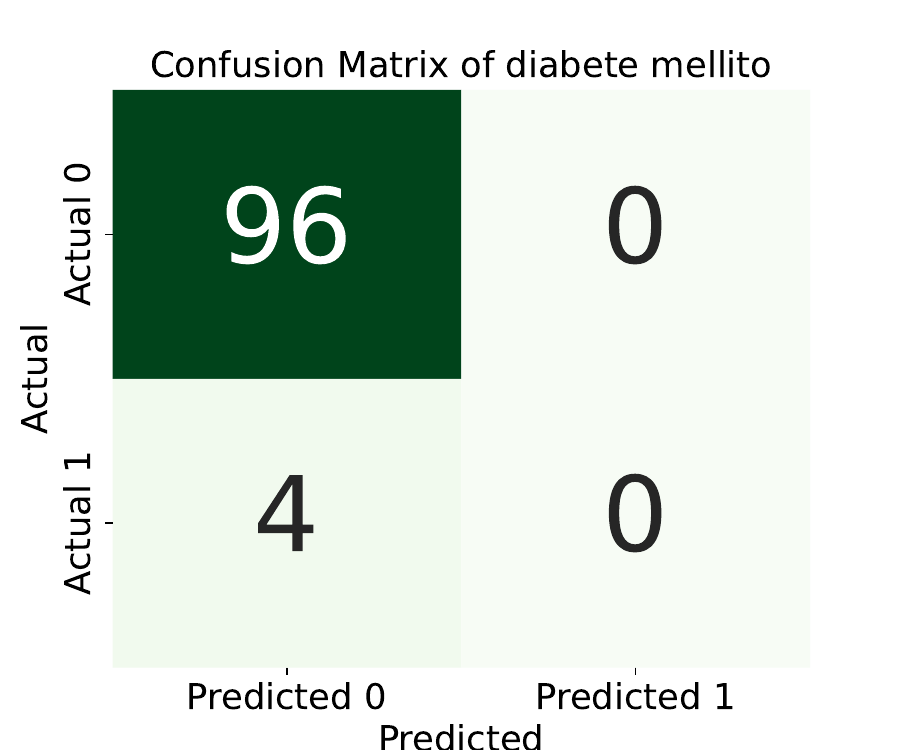}  
        \caption{}
    \end{subfigure}
    \begin{subfigure}[b]{0.19\textwidth}
        \includegraphics[width=\linewidth]{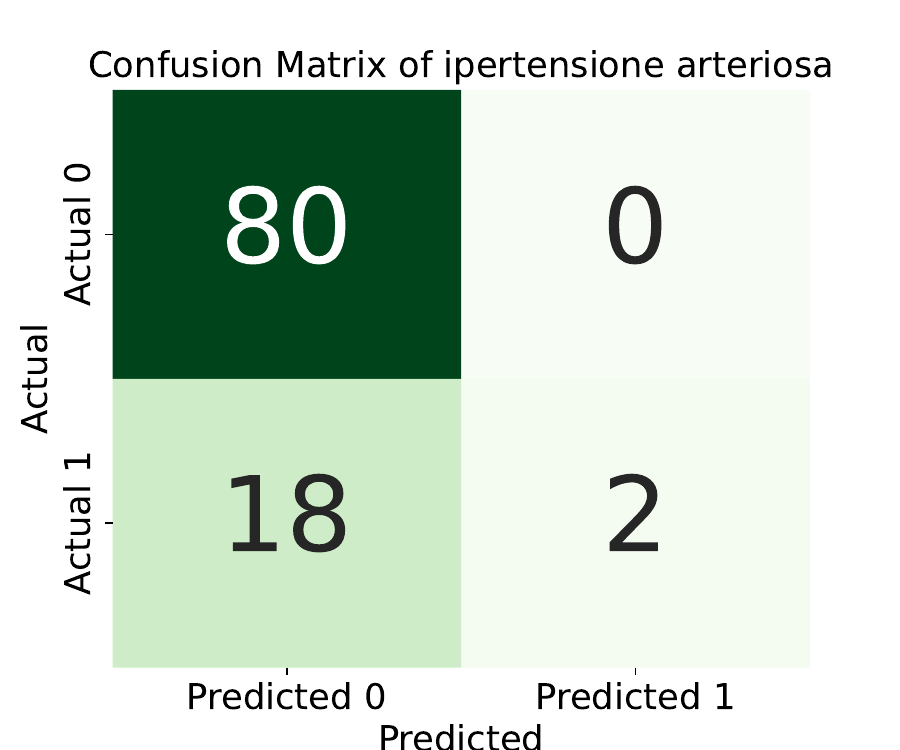}  
        \caption{}
    \end{subfigure}
    \end{subfigure}
    \caption{Confusion matrix for LLMs when compared to manual annotation. The first row represents OpenLLaMA 3B and the second, Mistral 7B. Comorbidities are (a,f) Fibrillazione atriale, (b,g) Insufficienza Renale, (c,h) BPCO-Broncopneumopatia cronica ostruttiva, (d,i) Diabete mellito and (e,j) Ipertensione arteriosa.}\label{fig:man_llama}
\end{figure}

From b) of the Figure~\ref{fig:llm_both} compared to the manual annotation, we could see that OpenLlama 3B shows a very low overall performance among all the models with a overall accuracy of less than 10~\% when compared to the manual annotation. OpenLlama 7B, Mistral 7B, Mixtral 8x7B, Qwen2.5 3B and 7B showed a overall accuracy of more than 80~\%. With mistral 7B and openllama 7B we see an accuracy increase of about $\approx$ 10~\% when compared to the automated annotation, Qwen2.5 3B and 7B we see an accuracy increase of about $\approx$ 20~\% when compared to the automated annotation and Mixtral 8x7B we see a whooping 47.48~\% of accuracy increase in the manual annotation when compared to the automated annotation. Regular expression had a accuracy of 92~\% when compared to manual annotation, llama and mixtral 8x7B are below that and the others three reports a similar performance that of regexp. In consideration of the huge unbalanced class distribution post the manual annotation, we use confusion matrix to compare the results with the LLMs instead of classification report. From b) of the Figure~\ref{fig:llm_both} we take the best and least accuracy model and  matrix is created and shown in Figure~\ref{fig:man_llama}. This is done to evaluate whether the model is generalizing effectively or simply making random predictions. Openllama 3B being we could clearly see a pattern that the model always classifies the comorbidities as positives on gathered subset of data for manual annotation. Mistral 7B shows a pattern, by classifying most of comorbidities as negatives on the gathered subset. This shows how the LLMs classify the selected comorbidities from the EHRs, showcasing a NO better performance compared to the regular expression approach. 

\begin{tcolorbox}
[width=\linewidth, sharp corners=all, colback=white!95!black]
Even though we see varied performance of LLMs across the comorbidities in this setting, we could conclude that Mistral 7B seems to generalize considerably well across three out of five comorbidities compared to the other LLMs when regular expression is considered. However, when direct comparison with manual annotation is considered the model always predicts class 0 (the majority class), thus not showing real understanding on the semantics of the comorbidities (\textbf{Q3}). 
\end{tcolorbox} 

\section{Discussion}

The results of the scenario~\ref{q1} considered in our study, and the extended technical evaluation of the same, provide valuable insights into the usage of LLMs in zero-shot, on-premises settings for comorbidities extraction from Italian EHRs. The initial accuracy results of the LLMs when compared to the regular expression or the manual annotation seem to be good, but when a classification report and confusion matrix are constructed and analyzed, they show how these LLMs struggle to generalize across all the chosen comorbidities.

The results from our evaluation demonstrate that Mistral 7B can achieve performance in extracting a few comorbidities from EHRs among the selected LLMs when regular expression is considered for comparison, but cannot generalize across all the chosen comorbidities in accordance with manual annotation. Adding to this, neither of the chosen LLMs could reach the level of the regular expression-based pattern matching approach. It is evident that the selected multilingual LLMs face serious challenges and trustworthiness related issues when used in a zero-shot setting with no technical expertise in high-risk domains as Healthcare. In this state of usage, considering LLMs in place of a pattern-matching approach in extraction pipelines would be problematic, since it would create uncertainty in the extracted information.

In context with the proposed scenario, when a clinician without much technical knowledge of prompt tuning techniques adheres to a simple and direct LLM prompt (zero-shot), aiming to extract comorbidities from EHRs, the overall results show that this approach would not be advisable in this setting. Although different prompt engineering approaches~\cite{sahoo2025systematicsurveypromptengineering} could be employed to increase the accuracy of LLMs usage and making them more adaptive to the task intended, we on purpose did not attempt to optimize or engineer the use of LLMs, thus replicating a simulated environment, where a clinician or doctor would be expected to use these models directly.

Overall, we want to highlight that the increasing use of LLMs without technical understanding in high-risk domains like healthcare, law, finance, and autonomous systems etc., brings concern. In these high-stakes environments, misinformation, biased decision-making, lack of accountability and security vulnerabilities are significant risks when deploying LLMs in critical fields, as mistakes could lead to dangerous outcomes. These analyses showed that LLMs must be properly tested before deployment into production, global metrics should be accurately selected to avoid misleading conclusions and continuous close monitoring should be done for hallucinations and other deviations.

\section{Conclusion}
In this paper, we explored the potential of six multilingual general-purpose open-source large language models (LLMs) in understanding and extracting valuable information from Electronic Health Records (EHRs) in Italian. Our study, focused on the real-time retrieval of comorbidities from EHRs, shows important insights into the LLMs ability to handle non-English language understanding and processing. Our findings show how LLMs struggle in extracting comorbidities and their difficulty in understanding Italian language EHR when used in a zero-shot approach. LLMs in the zero-shot setting cannot be used as a substitute for traditional pattern matching in the extraction pipelines. In the future, we will explore In-Context learning (ICL) approaches and will also consider fine-tuning an LLM model for IR from EHRs to improve its language processing capabilities and enhance trustworthiness in handling healthcare-related tasks.






% LNCS Homepage, \url{http://www.springer.com/lncs}, last accessed 2023/10/25
% \end{thebibliography}
\end{document}